\documentclass[10pt,twocolumn,letterpaper]{article}

\usepackage{utils/cvpr} 

\usepackage{xcolor}

\usepackage[normalem]{ulem}

\definecolor{cvprblue}{rgb}{0.21,0.49,0.74}
\usepackage[pagebackref,breaklinks,colorlinks,allcolors=cvprblue]{hyperref}
\usepackage{multirow}
\usepackage{amsmath}
\usepackage{amsfonts}
\usepackage{geometry}
\usepackage{booktabs}
\usepackage{array}
\usepackage{colortbl}
\usepackage{pifont}

\usepackage{xspace}
\usepackage{xcolor}

\definecolor{dsf_orange}{HTML}{F26A21} % 橙
\definecolor{dsf_pink}{HTML}{F57C8E}   % 粉
\definecolor{dsf_purple}{HTML}{A84D96} % 紫红

\newcommand{\ours}{%
\textcolor{dsf_pink}{G}%
\textcolor{dsf_pink!80!dsf_purple}{l}%
\textcolor{dsf_pink!60!dsf_purple}{a}%
\textcolor{dsf_pink!40!dsf_purple}{n}%
\textcolor{dsf_pink!20!dsf_purple}{c}%
\textcolor{dsf_purple}{e}\xspace
}

\usepackage[table]{xcolor} % 表格着色功能
\NewDocumentCommand{\inlineimage}{O{0.5} m}{%
  \raisebox{-0.25\baselineskip}{\includegraphics[height=#1\baselineskip]{#2}}\hspace{-1pt}
}

\newcommand{\icon}{\raisebox{0\height}{\inlineimage[1.]{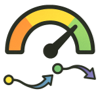}}}

\def\paperID{4562} % *** Enter the Paper ID here
\def\confName{CVPR}
\def\confYear{2026}

\newcommand{\tensamples}{%
  {\textcolor{dsf_pink}{1}%
   \textcolor{dsf_pink}{\,S}%
   \textcolor{dsf_pink!80!dsf_purple}{a}%
   \textcolor{dsf_pink!60!dsf_purple}{m}%
   \textcolor{dsf_pink!40!dsf_purple}{p}%
   \textcolor{dsf_pink!20!dsf_purple}{l}%
   \textcolor{dsf_purple}{e}}%
}

\title{\icon \ours: Accelerating Diffusion Models with \tensamples}

\newcommand{\name}{Slow-Fast}
\author{
Zhuobai Dong$^{1*}$,
Rui Zhao$^{2*}$, 
Songjie Wu$^{3}$, 
Junchao Yi$^{4}$,
Linjie Li$^{5}$,\\ 
Zhengyuan Yang$^{5}$, Lijuan Wang$^{5}$, 
Alex Jinpeng Wang$^{3}$
 \\[6pt]
$^{1}${WHU} \quad
$^{2}${NUS} \quad
$^{3}${CSU} \quad
$^{4}${UESTC} \quad
$^{5}${Microsoft} \\[6pt]
\url{https://zhuobaidong.github.io/Glance/}
}

\begin{document}
\pagestyle{plain}
\addtocontents{toc}{\protect\setcounter{tocdepth}{-1}}
\maketitle
\thispagestyle{plain}

\begin{abstract}

Diffusion models have achieved remarkable success in image generation, yet their deployment remains constrained by the heavy computational cost and the need for numerous inference steps. 
Previous efforts on fewer-step distillation attempt to skip redundant steps by training compact student models, yet they often suffer from heavy retraining costs and degraded generalization.
In this work, we take a different perspective: we accelerate smartly, not evenly, applying smaller speedups to early semantic stages and larger ones to later redundant phases.
We instantiate this phase-aware strategy with two experts that specialize in slow and fast denoising phases. 
Surprisingly, instead of investing massive effort in retraining student models, we find that simply equipping the base model with lightweight LoRA adapters achieves both efficient acceleration and strong generalization.
We refer to these two adapters as Slow-LoRA and Fast-LoRA.
Through extensive experiments, our method achieves up to \textbf{5$\times$ acceleration} over the base model while maintaining comparable visual quality across diverse benchmarks.
Remarkably, the LoRA experts are trained with only 1 samples on a single V100 within one hour, yet the resulting models generalize strongly on unseen prompts.
%Project page: \href{https://zhuobaidong.github.io/Glance/}{Glance.github.io}

%
% Even more intriguingly, our approach achieves such acceleration with an extremely lightweight training setup: each LoRA expert is trained using only 10 data samples on a single V100 GPU within five hours—a surprisingly small cost that challenges conventional beliefs about diffusion distillation. 
%
% Despite this minimal supervision, the resulting models exhibit remarkable generalization ability, matching the teacher’s visual fidelity while preserving strong adaptability across unseen prompts. 
% %
% On FLUX.1-12B and Qwen-Image-20B at 8 NFEs, our method consistently produces richer diversity and teacher-level quality, revealing a new perspective on efficient diffusion acceleration through phase-aware LoRA design.
% 
% \rui{do we really need to mention 8 NFEs or 4 NFEs}

\end{abstract}

\footnotetext[1]{* Equal contribution.}

\section{Introduction}
\label{sec:intro}

\begin{figure}[t]  % 去掉星号
  \centering
  \includegraphics[width=\columnwidth]
  {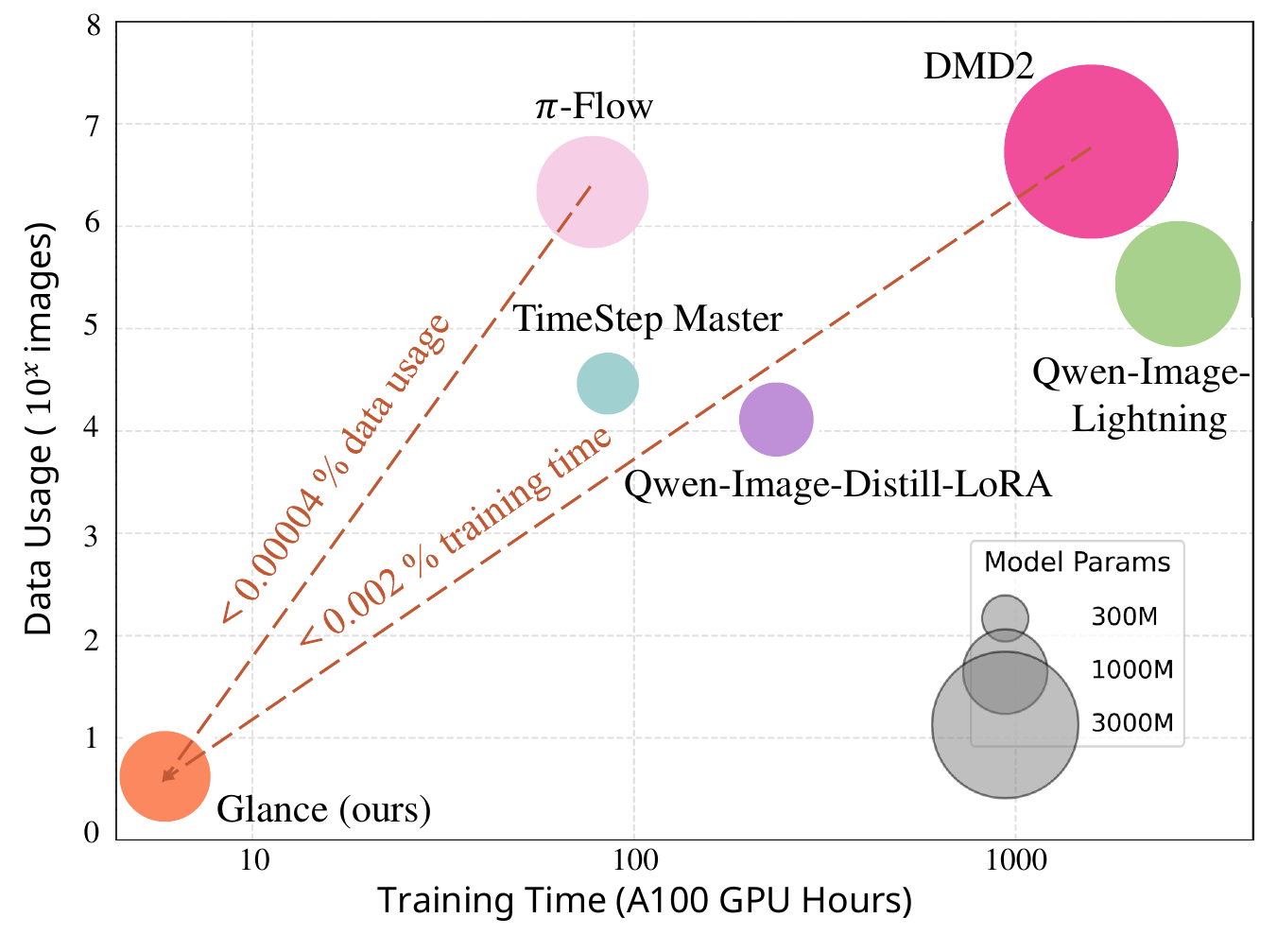}  % 使用 \linewidth 而不是 \textwidth
  \caption{
  \textbf{Comparison of data usage and training time.} 
  \ours achieves comparable generation quality with only \textbf{1 training samples and within 1 GPU-hour}, demonstrating extreme data and compute efficiency. Note that the x-axis is in logarithmic scale, and values equal to zero are therefore not representable.
  }
  \label{fig:introduction}
  % \vspace{-1em}
\end{figure}

\begin{figure*}[t]
  \centering
   \includegraphics[width=1\textwidth]{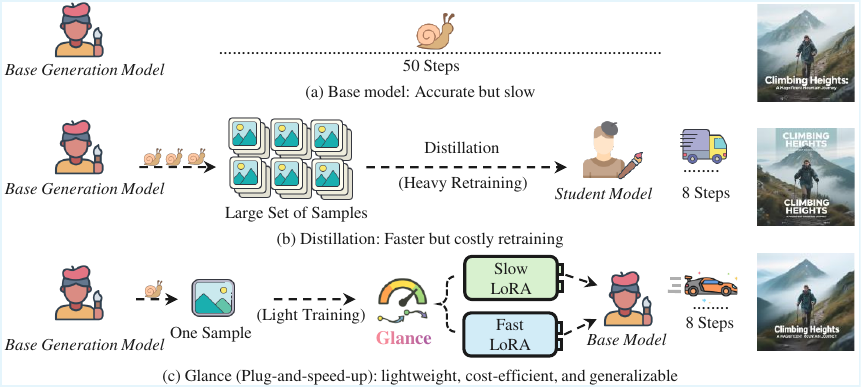}
   \caption{
   \textbf{Comparison of distill and accelerate strategies}. 
   % We find that different denoising steps have profoundly different influence on high and low-level features of the resulting images.
   Prior distillation pipelines rely on large training sets and costly retraining. \ours requires only one training sample to obtain Slow-LoRA and Fast-LoRA, providing plug-and-play acceleration of the base generation model.
  }
   \label{fig1}
   
\vspace{-1em}
\end{figure*}

%%%% bg, raise of distilled model
Diffusion and flow matching models \cite{sohl2015deep,ho2020denoising,song2019generative,lipman2022flow,albergo2022building} have shown strong capabilities in generating high-fidelity images, marking a significant advancement in the field of generative modeling. Despite their impressive performance, a notable challenge is the high inference cost due to its iterative denoising nature. 
To address this issue, various methods are proposed to accelerate the sampling process of diffusion models, including improving the efficiency of samplers \cite{karras2022elucidating,lu2022dpm,lu2022dpm_p,liu2022pseudo} and employing model distillation techniques \cite{salimans2022progressive,liu2022flow,frans2024one,song2023consistency,kim2023consistency,geng2025mean} to reduce the number of inference steps. 
Recent advancements in trajectory distillation methods and distribution matching techniques \cite{yin2024one,yin2024improved,sauer2024adversarial,zhou2024score}, often enhanced by adversarial learning at scale, have shown considerable promise in generating high-fidelity images in extremely low steps such as one to four steps.

Despite significant advancements in timestep-distilled diffusion models, it remains unclear how to effectively fine-tune or customize such distilled models. 
Naively tuning the distilled model with diffusion loss will make the generation results blurry. 
An alternative approach is to fine-tune or customize the original diffusion model, and then repeat the diffusion distillation process to create a distilled model variant. 
However, the large computation cost of diffusion distillation, when compared with the customization training used for distillation (cf., 3840 A100 GPU hours for SDXL-DMD2 \cite{yin2024improved} and 3072 A100 GPU hours for Qwen-Image-Lightning \cite{ModelTC2025QwenImageLightning}), often makes such distilled model tuning approach less feasible. 

In this work, we take a different perspective by revisiting the denoising dynamics of diffusion models. 
We observe that the generation trajectory consists of two qualitatively distinct phases: an early semantic phase that determines global structure, and a late redundant phase that primarily refines texture. 
Uniform acceleration treats all steps equally, yet semantic steps are far more sensitive to perturbation than redundant ones. 
This motivates a phase-aware acceleration strategy that applies small speedups to semantic steps and large speedups to redundant steps.

To realize this idea, we introduce \ours, implemented as a pair of lightweight LoRA adapters that attach to the pretrained diffusion model. 
Slow-LoRA stabilizes early semantic formation, while Fast-LoRA accelerates late-stage refinement. 
Crucially, our method does not require training a new student network and the base model remains unchanged. 
Both LoRA experts are trained using only \textbf{one sample} on a single V100 GPU within \textbf{one hour} (Fig.~\ref{fig:introduction}). 

To demonstrate its scalability, we distill FLUX.1-12B \cite{flux2024} and Qwen-Image-20B \cite{wu2025qwenimagetechnicalreport} text-to-image models into 8- and 10-step students, respectively.
Extensive experiments across six text-to-image benchmarks show that \ours exhibits performance curves that closely track those of the base models, indicating strong consistency under accelerated inference. 
On OneIG-Bench, HPSv2, and GenEval, the performance of \ours reaches \textbf{92.60\%}, \textbf{99.67\%}, and \textbf{96.71\%} of the teacher models, respectively. 
We further conduct dense ablation studies on slow–fast phase decomposition, timestep allocation, and training data scaling, all of which consistently validate the effectiveness and robustness of our design.

We summarize the contributions as follows: 
% % (1) We employ a phase-aware acceleration scheme that treats semantic and redundant steps differently for a more natural and stable speedup.
% (2) We introduce Slow-LoRA and Fast-LoRA, two lightweight adapters that plug into the base model, enabling effective phase-wise acceleration with only one sample and one hour of training.
%
% (3) \ours achieves a \textbf{5$\times$ speed-up} over FLUX.1 and Qwen-Image while retaining teacher-level performance across six benchmarks.

\begin{itemize}

    \item 
    We employ a phase-aware acceleration scheme that treats semantic and redundant steps differently for a more natural and stable speedup.
    
    \item 
    We introduce two lightweight adapters that plug into the base model, enabling effective acceleration with only one sample and one hour of training.

    \item 
    \ours achieves a \textbf{5$\times$ speed-up} over FLUX.1 and Qwen-Image while retaining teacher-level performance across six benchmarks.

\end{itemize}

\begin{figure*}[t]
  \centering

    \vspace{-1em}
   \includegraphics[width=\textwidth]{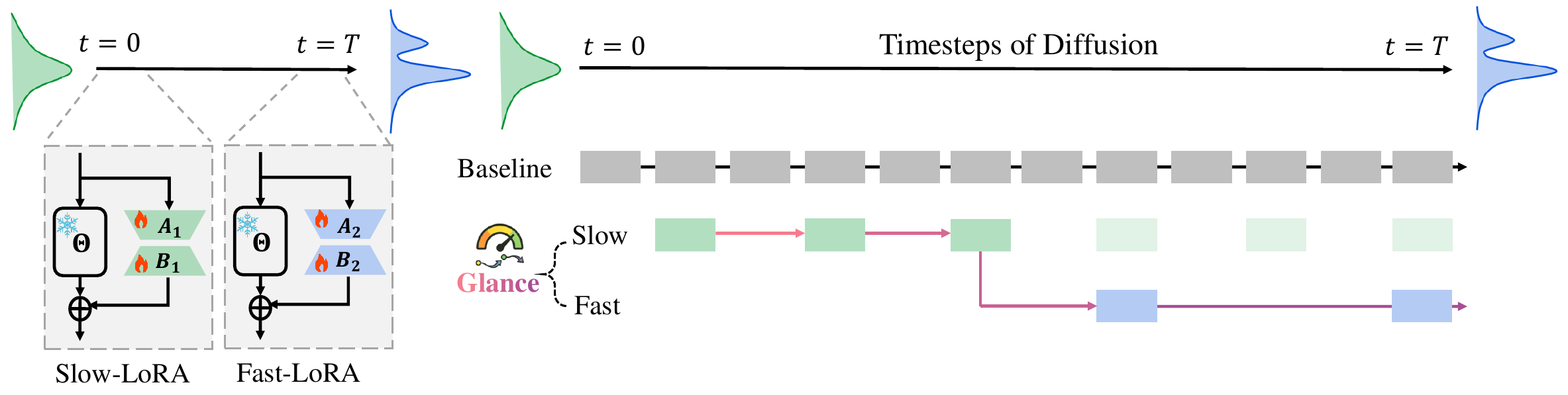}
   \caption{\textbf{Visualization of Slow-Fast paradigm}. 
   In the slow stage, we sample one timestep every two steps from the first 20 timesteps (i.e., 5 samples in total).
   In the fast stage, an additional 5 timesteps are uniformly sampled from the remaining 40 steps. During inference, the slow-stage timesteps are executed prior to the fast-stage ones.}
   \label{fig:onecol}
   
\vspace{-1em}
\end{figure*}

% \begin{figure*}
%   \centering
%   \begin{subfigure}{0.68\linewidth}
%     \fbox{\rule{0pt}{2in} \rule{.9\linewidth}{0pt}}
%     \caption{An example of a subfigure.}
%     \label{fig:short-a}
%   \end{subfigure}
%   \hfill
%   \begin{subfigure}{0.28\linewidth}
%     \fbox{\rule{0pt}{2in} \rule{.9\linewidth}{0pt}}
%     \caption{Another example of a subfigure.}
%     \label{fig:short-b}
%   \end{subfigure}
%   \caption{Example of a short caption, which should be centered.}
%   \label{fig:short}
% \end{figure*}

\section{Related Work}
\label{sec:formatting}

\paragraph{Diffusion Models.} 

Recently, diffusion models (DMs) \cite{sohl2015deep,ho2020denoising,song2020score} have become the leading paradigm for visual generation, achieving state-of-the-art performance across a wide range of conditioning modalities, including images \cite{rombach2022high}, depth, edges, poses \cite{zhang2023controlnet,mou2024t2i}, and text \cite{dhariwal2021diffusion,ramesh2022hierarchical,rombach2022high,esser2024scaling,ho2022imagenvid,podell2023sdxl}.
Advances in large-scale systems such as PixArt \cite{chen2023pixart-alpha,chen2024pixart-sigma,chen2024pixart-delta}, SD3 \cite{esser2024sd3}, Qwen-Image \cite{wu2025qwenimagetechnicalreport}, and FLUX \cite{flux2024} further push generation quality, controllability, and multilingual rendering capabilities.
Despite these rapid developments, achieving high-fidelity synthesis still requires many denoising steps, resulting in substantial inference cost and limiting their use in real-time or resource-constrained applications.

\paragraph{Diffusion Distillation.}
Early work \cite{luhman2021knowledge} directly regresses the teacher’s ODE integral in a single step, but the $\ell_2$-based $x_0$ regression often produces overly smooth and blurry results. 
Progressive distillation methods \cite{salimans2022progressive,liu2022flow,frans2024one} refine this paradigm via multi-stage training that enlarges step size and lowers NFE by merging teacher steps.
While effective, these approaches suffer from error accumulation and substantial computational overhead. Consistency distillation \cite{song2023consistency,kim2023consistency,geng2025mean} replaces $x_0$ regression with velocity-based objectives to enforce trajectory consistency, improving fidelity but requiring costly Jacobian–vector products (JVPs) or inaccurate finite-difference approximations. Distribution matching methods \cite{yin2024one,yin2024improved,sauer2024adversarial,zhou2024score}instead adopt score-based or adversarial objectives to align the student’s output distribution with the teacher’s, achieving high perceptual quality yet prone to mode collapse and instability due to auxiliary discriminators. 
Recent concurrent studies \cite{chen2025pi,DiffSynthStudio2025,ModelTC2025QwenImageLightning} have distilled Qwen-Image and FLUX into compact 4-NFE or 8-NFE versions, but these still incur high distillation costs from trajectory-level supervision and extensive teacher sampling. In contrast, our approach achieves comparable or superior generation quality with minimal training cost, avoiding recursive distillation and auxiliary network overhead while maintaining stable optimization.

% \paragraph{Efficient Tuning of Diffusion Models.}
% To reduce the cost of full fine-tuning while preserving generalization, LoRA \cite{hu2022lora} has been widely adopted in diffusion models for efficient adaptation through low-rank parameter updates \citep{zhang2023controlnet,ye2023ipadapter,xie2023difffit,mou2024t2iadapter,lin2024ctrladapter,xing2024simda,ran2024x,gu2024mix,lyu2024one,huang2023t2i}. Building on this, several studies integrate LoRA into distillation or controllable generation frameworks. DMD \citep{yin2024dmd} enables LoRA-based model distillation for faster inference, while ControlNeXt \citep{peng2024controlnext} leverages LoRA for enhanced controllability. TC-LoRA~\cite{cho2025tc} performs dynamic weight distillation by using a hypernetwork to generate time- and condition-dependent LoRA adapters at each diffusion step, enabling adaptive control throughout generation. Timestep Master \cite{zhuang2025timestep} assigns LoRA experts to different noise levels for improved representation aggregation. Recent efforts such as Qwen-Image-Lightning \cite{ModelTC2025QwenImageLightning} and Qwen-Image-Distill-LoRA \cite{DiffSynthStudio2025} further embed LoRA into distilled diffusion backbones, producing compact 4-NFE or 8-NFE models. 
% However, the generation capabilities of LoRA-tuned DMs are limited. 
% We tackle this by inserting LoRA adapters more smartly, allocating slow and fast adapters according to denoising phases, which in turn yields substantial improvements.

\paragraph{Efficient Tuning of Diffusion Models.}  
LoRA \cite{hu2022lora} has been widely adopted for efficient adaptation of diffusion models via low-rank parameter updates \citep{zhang2023controlnet,ye2023ipadapter,xie2023difffit,mou2024t2iadapter,lin2024ctrladapter,xing2024simda,ran2024x,gu2024mix,lyu2024one,huang2023t2i}. Several works integrate LoRA into distillation or controllable generation: DMD \citep{yin2024dmd} enables faster inference through LoRA-based distillation; ControlNeXt \citep{peng2024controlnext} and TC-LoRA \citep{cho2025tc} provide adaptive, time- or condition-dependent LoRA control; Timestep Master \citep{zhuang2025timestep} assigns LoRA experts to different noise levels for better representation. Recent models such as Qwen-Image-Lightning \cite{ModelTC2025QwenImageLightning} and Qwen-Image-Distill-LoRA \cite{DiffSynthStudio2025} embed LoRA into distilled backbones, producing compact low-step (4–8 NFE) models. However, their generation quality remains limited. We address this by allocating slow and fast LoRA adapters according to denoising phases, yielding substantial improvements.

\section{Method}
\label{sec:method}

% In this section, we introduce our slowfast paradigm designed to enhance the efficiency and adaptability of diffusion models through phase-aware low-rank adaptation.  
% We first review the diffusion model and LoRA as preliminaries, and then elaborate on our two-stage design that builds a mixture of TimeStep LoRA experts for efficient and versatile denoising enhancement.

In this section, we introduce~\ours, a phase-aware acceleration framework that improves both efficiency and adaptability of diffusion models through slow–fast paradigm. We first revisit the diffusion model and flow-matching formulation as preliminaries, then describe our phase-aware LoRA experts and their learning objectives.

\subsection{Preliminary}
\paragraph{Diffusion and Flow Matching.}

Diffusion models~\cite{ho2020denoising} learn data distributions by gradually transforming noise into data through a parameterized denoising process. The  flow matching formulation~\cite{liu2022flow,esser2024scaling} interprets diffusion as learning a continuous velocity field that transports a sample from Gaussian noise $x_1 \!\sim\! \mathcal{N}(0, I)$ to clean data $x_0$.
At timestep $t\!\in\![0,1]$, the intermediate state is defined as $x_t = t x_0 + (1 - t)x_1$, and the model predicts the transport velocity $v_\theta(x_t, t, h)$ conditioned on guidance $h$ (e.g., text embedding).
The objective is a mean-squared error between the predicted and target velocities:
\begin{equation*}
\mathcal{L}_{\text{FM}} =
\mathbb{E}_{x_0, x_1, t, h}
\!\left[
\|\, v_\theta(x_t, t, h) - v_t \,\|_2^2
\right],
\label{eq:fm_loss}
\end{equation*}
where $v_t$ is the groundtruth velocity. 
% This formulation yields stable ODE-based training and aligns with the maximum-likelihood objective under mild assumptions.
%
To achieve superior performance,
the diffusion model is often designed with a large number of network parameters that are pre-trained on large-scale web data.
Apparently,
it is computationally expensive to distill such a big model for step reduction.

\paragraph{Low-Rank Adaptation}

To alleviate the above difficulty, LoRA \cite{hu2022lora} has been recently applied for rapid distillation diffusion models on target data \cite{zhuang2025timestep,chadebec2025flash}. Specifically, LoRA introduces low-rank decomposition of an extra matrix,
% \begin{equation}
% \Theta + \Delta\Theta = \Theta + BA,
% \label{eq:lora}
% \end{equation}
$\Theta' = \Theta + B A,$
where $\Theta \in \mathbb{R}^{d\times k}$ denotes the frozen pretrained parameters, and the low-rank matrices $A \in \mathbb{R}^{r\times k}$ and $B \in \mathbb{R}^{d\times r}$ (with $r \ll d,k$) constitute the learnable LoRA parameters.  

% This design significantly reduces the number of trainable parameters and allows rapid distillation.  
% However, existing LoRA-based diffusion models employ the same pair of low-rank matrices $(A, B)$ across all timesteps, ignoring that the denoising dynamics vary substantially across different noise levels. 
 
% Such a mismatch limits the model’s capacity to handle heterogeneous noise regimes effectively.

% To overcome this limitation, we propose the slow-fast paradigm, which introduces \emph{phase-aware LoRA experts} to specialize in different denoising phases of the diffusion process.

\subsection{Phase-aware LoRA Experts for Phase-wise Denoising}
\label{sec:phase_lora}

\begin{table*}[t]
    \centering
    \caption{
    \textbf{Quantitative comparisons on COCO-10k dataset and HPSv2 prompt set.}}
    \label{tab:fluxqwen}
    \scalebox{0.89}{%
        \setlength{\tabcolsep}{0.35em}
        \begin{tabular}{lccccccccccc}
        \toprule
        \multirow{3}[3]{*}{\textbf{Model}} & \multirow{3}[3]{*}{\textbf{Distill method}} & \multirow{3}[3]{*}{\textbf{NFE}} & \multicolumn{5}{c}{COCO-10k prompts} & \multicolumn{3}{c}{HPSv2 prompts} \\
        \cmidrule(lr){4-8} \cmidrule(lr){9-11} 
        & & & \multicolumn{2}{c}{Data align.} & \multicolumn{2}{c}{Prompt align.} & Pref. align. & \multicolumn{2}{c}{Prompt align.} & Pref. align. \\
        \cmidrule(lr){4-5} \cmidrule(lr){6-7} \cmidrule(lr){8-8} \cmidrule(lr){9-10} \cmidrule(lr){11-11}
        & & & \textbf{FID\textdownarrow} & \textbf{pFID\textdownarrow} & \textbf{CLIP\textuparrow} & \textbf{VQA\textuparrow} & \textbf{HPSv2.1\textuparrow} & \textbf{CLIP\textuparrow} & \textbf{VQA\textuparrow} & \textbf{HPSv2.1\textuparrow} \\
        \midrule
        FLUX.1 dev
            & - & 50 & 27.8 & 34.9 & 0.268 & 0.900 & 0.309 & 0.284 & 0.805 & 0.314 \\
        \midrule
        FLUX Turbo 
            & GAN & 8 & \textbf{26.7} & \textbf{32.0} & 0.267 & 0.900 & 0.308 & \textbf{0.286} & \textbf{0.814} & 0.313 \\
        Hyper-FLUX 
            & CD+Re & 8 & 29.8 & 33.3 & \textbf{0.268} & 0.894 & 0.309 & 0.285 & 0.807 & 0.315 \\
        $\pi$-Flow (FLUX)
            & $\pi$-ID & 8 & 29.0 & 35.4 & \textbf{0.268} & \textbf{0.901} & \textbf{0.311} & 0.285 & 0.810 & \textbf{0.316} 
            \\
        \textbf{\ours{} (FLUX)}
            & Slow-Fast & 8 & 34.2 & 40.1 & 0.259 & 0.879 & 0.298 & 0.276 & 0.895 & 0.297 \\
        \textbf{\ours{} (FLUX)}
            & Slow-Fast & 10 & 30.4 & 37.5 & 0.265 & 0.891 & 0.303 & 0.282 & 0.799 & 0.303 
            \\
        \midrule
        Qwen-Image 
            & - & 50\texttimes2 & 34.1 & 45.6 & 0.282 & 0.936 & 0.312 & 0.302 & 0.872 & 0.309 \\
        Qwen-Image 
            & - & 10\texttimes2 & 39.1 & 52.1 & 0.265 & 0.918 & 0.287 & 0.288 & 0.835 & 0.305 \\
        Qwen-Image 
            & - & 8\texttimes2 & 44.2 & 56.3 & 0.263 & 0.873 & 0.269 & 0.287 & 0.761 & 0.292 \\
        \midrule
        Qwen-Image-Lightning 
            & VSD & 4 & 37.5 & 51.6 & 0.280 & \textbf{0.935} & \textbf{0.322} & 0.299 & \textbf{0.867} & \textbf{0.328} \\
        $\pi$-Flow (Qwen)
            & $\pi$-ID & 4 & \textbf{36.0} & \textbf{46.1} & \textbf{0.281} & 0.934 & 0.314 & \textbf{0.300} & 0.860 & 0.310 \\
        % \textbf{\ours{} (Qwen)} 
        %     & Timestep & 8\texttimes2 & 38.9 & 52.4 & 0.273 & 0.926 & 0.289 & 0.287 & 0.842 & 0.302  \\
        % \textbf{\ours{} (Qwen)} 
        %     & Timestep & 10\texttimes2 & 37.2 & 49.6 & 0.279 & 0.930 & 0.308 & 0.299 & 0.851 & 0.311 
        %     \\
        % \textbf{\ours{} (Qwen)} 
        %     & Timestep & 15\texttimes2 & * & * & * & * & * & 0.300 & 0.857 & 0.332 
        %     \\
        \textbf{\ours{} (Qwen)} 
            & Slow-Fast & 8\texttimes2 & 38.9 & 52.4 & 0.273 & 0.926 & 0.289 & 0.287 & 0.842 & 0.302 
            \\
        \textbf{\ours{} (Qwen)} 
            & Slow-Fast & 10\texttimes2 & 37.8 & 50.3 & 0.280 & 0.932 & 0.313 & 0.297 & 0.849 & 0.308 
            \\
        \bottomrule
        \end{tabular}
    }
\end{table*}

\begin{table*}[t]
    \centering
    \caption{\textbf{Quantitative comparisons on OneIG-Bench.} * denotes unavailable results.}
    \label{tab:oneig}
    \scalebox{0.89}{%
        \setlength{\tabcolsep}{0.35em}
        \begin{tabular}{lccccccccc}
        \toprule
        \textbf{Model} & \textbf{Distill Method} & \multicolumn{2}{c}{\textbf{Training Cost}} & \textbf{NFE} & \textbf{Alignment\textuparrow} & \textbf{Text\textuparrow} & \textbf{Diversity\textuparrow} & \textbf{Style\textuparrow}  & \textbf{Reasoning\textuparrow} \\
        \cmidrule(lr){3-4}
        & & \textbf{Data} & \textbf{GPU hour} &  &  &  &  &  &  \\
        \midrule
        FLUX.1 dev 
            & - & - & - & 50 & 0.790 & 0.556 & 0.238 & 0.370 & 0.257 \\
        \midrule
        FLUX Turbo 
            & GAN & 1M & * & 8 & 0.791 & 0.334 & \textbf{0.234} & \textbf{0.370} & 0.239 \\
        Hyper-FLUX 
            & CD+Re & 1.1M & 800 & 8 & 0.790 & \textbf{0.530} & 0.198 & 0.369 & 0.254 \\
        $\pi$-Flow (FLUX)
            & $\pi$-ID & 2.3M & 83 & 8 & \textbf{0.792} & 0.517 & \textbf{0.234} & 0.369 & \textbf{0.256} \\
        \textbf{\ours{} (FLUX)} 
            & Slow-Fast & 1 & 0.6 & 8 & 0.774 & 0.284 & 0.196 & 0.353 & 0.208 \\
        \textbf{\ours{} (FLUX)} 
            & Slow-Fast & 1 & 0.8 & 10 & 0.788 & 0.328 & 0.204 & 0.358 & 0.231 \\
        \midrule
        Qwen-Image 
            & - & - & - & 50\texttimes2 & 0.880 & 0.888 & 0.194 & 0.427 & 0.306 \\
        Qwen-Image 
            & - & - & - & 10\texttimes2 & 0.802 & 0.693 & 0.156 & 0.410 & 0.290  \\
        Qwen-Image 
            & - & - & - & 8\texttimes2 & 0.752 & 0.611 & 0.148 & 0.411 & 0.276 \\
        \midrule
        Qwen-Image-Lightning & VSD & 0.4M & 3072 & 4 & \textbf{0.885} & \textbf{0.923} & 0.116 & 0.417 & \textbf{0.311} \\
        $\pi$-Flow (Qwen)
            & $\pi$-ID & 2.3M & 83 & 4 & 0.875 & 0.892 & \textbf{0.180} & \textbf{0.434} & 0.298 \\
        LoRA (Qwen)
            & Uniform steps & 1 & 0.8 & 10\texttimes2 & 0.621 & 0.332 & 0.097 & 0.298 & 0.193 \\
        \textbf{\ours{} (Qwen)} 
            & Slow-Fast & 1 & 0.6 & 8\texttimes2 & 0.863 & 0.692 & 0.162 & 0.414 & 0.286 \\
        \textbf{\ours{} (Qwen)} 
            & Slow-Fast & 1 & 0.8 & 10\texttimes2 & 0.868 & 0.734 & 0.160 & 0.421 & 0.303 \\
        \bottomrule
        \end{tabular}
    }
    
\vspace{-1em}
\end{table*}

To accelerate the denoising process of pretrained diffusion models while maintaining generative quality, 
we retain the pretrained parameters~$\Theta$ and introduce a compact yet effective augmentation: 
a set of \emph{phase-specific} LoRA adapters. 
Each adapter specializes in a specific stage of the denoising trajectory, enabling the model to adapt dynamically to varying noise levels and semantic complexities during inference.

\paragraph{Beyond uniform timestep partitioning.}
Prior works such as Timestep Master~\cite{zhuang2025timestep} 
have demonstrated the potential of using multiple LoRA adapters trained over different timestep intervals. 
However, Uniform partitioning assumes equal contribution from all timesteps, which contradicts the intrinsic non-uniformity of diffusion dynamics.
Empirical analyses and prior studies~\cite{anagnostidis2025flexidit} 
reveal that different timesteps exhibit markedly different levels of semantic importance: 
in the early, high-noise regime, the model primarily reconstructs coarse global structures and high-level semantics 
(\emph{low-frequency information}); 
in contrast, the later, low-noise regime refines textures and details (\emph{high-frequency information}). 
% Consequently, early steps play a disproportionately critical role in determining the perceptual and structural integrity of the final image, 
% while later steps mainly handle appearance refinement.

% Uniform splitting fails to respect this imbalance, leading to inefficient specialization:
% early-stage LoRAs are under-capacitated, while late-stage LoRAs tend to be underutilized.

\paragraph{Phase-aware partitioning via SNR.}
To better align expert specialization with the intrinsic dynamics of the diffusion process, 
we introduce a \emph{phase-aware} partitioning strategy guided by the signal-to-noise ratio (SNR). 
Unlike timestep indices, the SNR provides a physically meaningful measure of the relative dominance between signal and noise, 
and it decreases monotonically as denoising progresses. 
At the beginning of the process (\(t\) large, high-noise phase), the latent representation is dominated by noise with a low SNR, 
making coarse structural recovery the primary objective. 
In contrast, as \(t\) decreases and SNR rises, the model transitions into a low-noise regime focused on texture refinement.

Based on this observation, we define a transition boundary \(t_{\mathrm{s}}\) corresponding to an SNR threshold 
(e.g., half of the initial SNR value). 
Two phase-specific experts are then employed: 
a \emph{slow expert} specialized for the high-noise phase (\(t \ge t_{\mathrm{s}}\)) that focuses on coarse semantic reconstruction, 
and a \emph{fast expert} for the low-noise phase (\(t < t_{\mathrm{s}}\)) that enhances fine-grained details. 
This SNR-guided partition allows each expert to operate in the regime where it is most effective, 
forming a semantically meaningful decomposition of the denoising process.

\paragraph{Surprising effectiveness of extremely small training sets.}
To evaluate whether phase-wise LoRAs can recover accelerated inference, we initially conducted an overfitting-style experiment using only \textbf{10 training samples}.
Unexpectedly, the model rapidly learned a faithful approximation of the accelerated sampling trajectory.
Even more remarkably, reducing the dataset to \textbf{a single training sample} still produced a stable acceleration behavior.

We attribute this  data efficiency to the nature of flow matching. 
By directly predicting the target velocity field along the diffusion trajectory, the training objective bypasses redundant score-matching steps. 
Consequently, essential structural knowledge for fast inference can be distilled from only a few examples.

\paragraph{Necessity of carefully designed timestep skipping.}
Despite this promising data efficiency, subsequent ablation studies reveal that timestep skipping is far from arbitrary.
Although few-step students can imitate the teacher behavior in aggregate, not all timesteps contribute equally to the reconstruction dynamics; naive skipping strategies can severely degrade performance.

To this end, we conducted a comprehensive investigation of different specialization schemes.
We first explored assigning multiple timesteps to the slow stage LoRA adapters while keeping a single adapter for the fast stage, and vice versa.
We also tested a degenerate configuration where a single LoRA was trained across the entire trajectory.
However, these variants either lacked the expressiveness to capture high-noise complexity or failed to exploit temporal locality in the low-noise refinement phase.

Our experiments ultimately show that separating the trajectory into a dedicated slow region and a dedicated fast region yields the most robust specialization.
This design preserves sufficient capacity for modeling the challenging high-noise dynamics while enabling lightweight refinement in later steps, achieving a compact yet effective acceleration mechanism.

\paragraph{Flow-matching supervision.}
Each phase-specific LoRA expert is trained under a flow-matching supervision scheme that aligns its predicted denoising direction with the underlying data flow.  
Given the noisy latent $x_t$ obtained during the diffusion process, the model predicts a velocity field $\hat{v}_{\Theta,B_i,A_i}(x_t, t, c)$, which is supervised against the ground-truth flow vector $v_t^\star$.  
The training objective is defined as a weighted mean-squared error: 
\begin{equation*}
\mathcal{L}_{\mathrm{FM}}(t; i) =
\mathbb{E}_{x, c, \varepsilon}
\big[\, w(t)\,\|\hat{v}_{\Theta,B_i,A_i}(x_t, t, c) - v_t^\star\|_2^2 \,\big],
\label{eq:flow_loss}
\end{equation*}
where $w(t)$ denotes an optional timestep-dependent weighting function.  
By restricting the training samples of each expert to its assigned denoising phase, the model effectively learns to specialize on distinct noise levels.  
The resulting mixture of phase-aware LoRA experts collectively improves both the denoising speed and generative quality, forming the foundation of our proposed \emph{slowfast} paradigm.

\begin{figure}[t]
  \centering

   \includegraphics[width=0.48\textwidth]{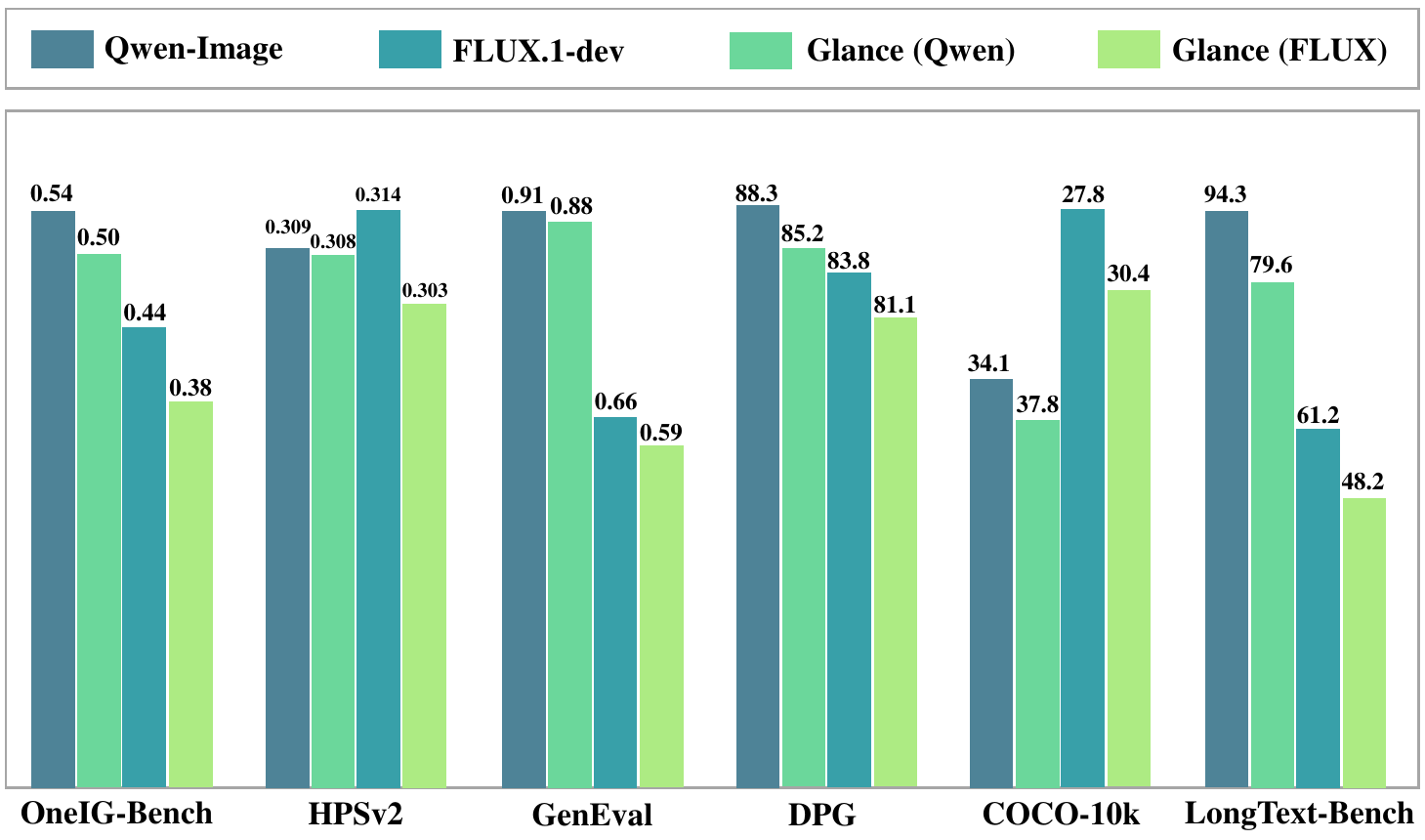}
   
\vspace{-1em}
   \caption{\textbf{Comparison between Image Generation Benchmarks}. Glance further shows performance trajectories that closely follow those of the corresponding base models.}
   \label{fig:all_bench}
   
\vspace{-1em}
\end{figure}
\begin{figure*}[t]
  \centering
   \includegraphics[width=\textwidth]{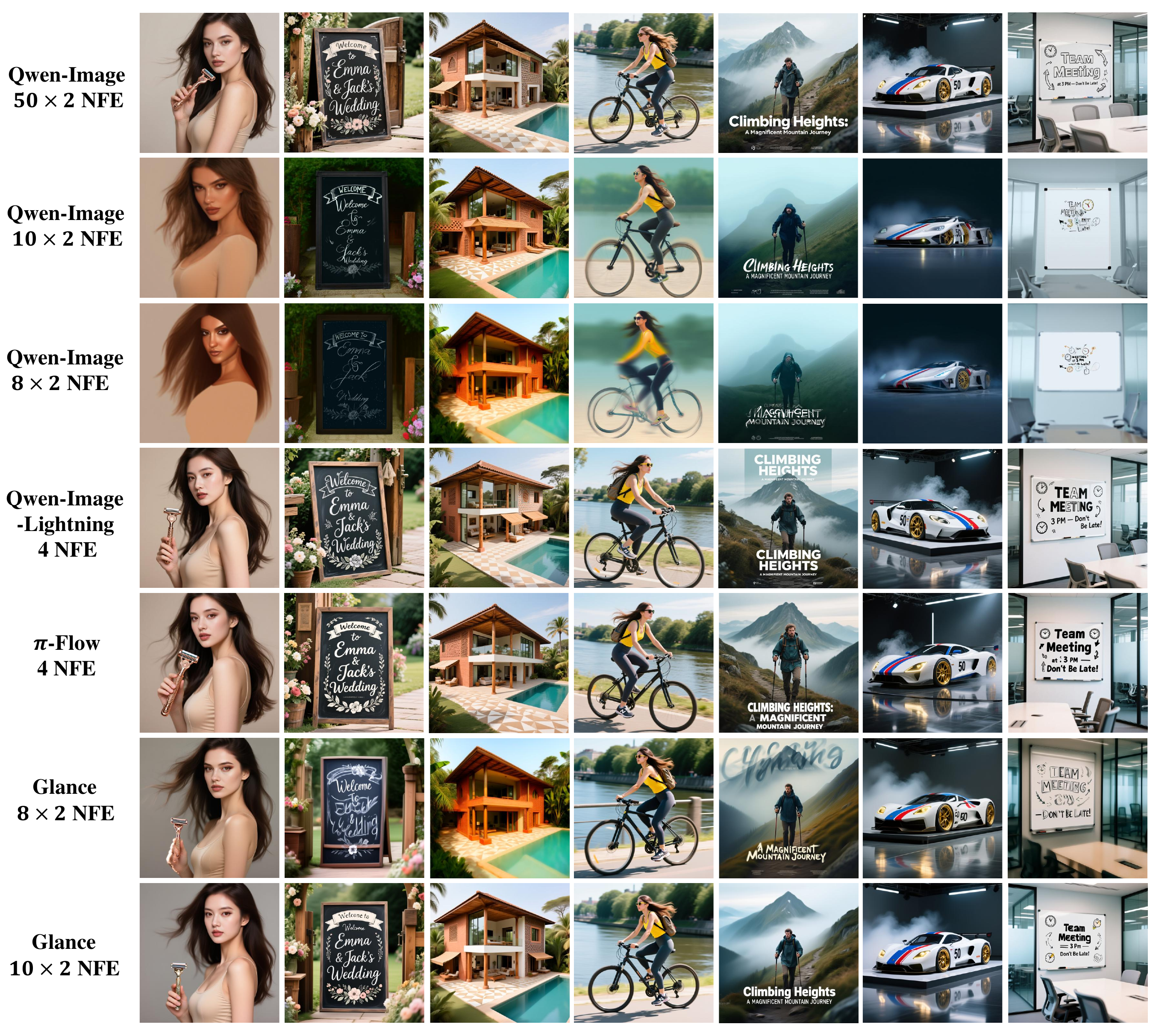}
   
\vspace{-1em}
   \caption{
    \textbf{Visual comparison of different Slow–Fast configurations.}
    All images are generated from the same initial noise using the 50-step base model, our 8/10-step students, and other few-step models.
    \name{} preserves semantic fidelity under strong acceleration, while additional steps progressively enhance fine details.
   }
   \label{fig:compare_with_base_model}
   
\vspace{-1em}
\end{figure*}
\section{Experiments}
\label{sec:exp}

This section presents a comprehensive evaluation of \ours on the text-to-image generation task. 
We first report quantitative results compared with competitive baselines, followed by detailed ablation analyses. 
We then discuss the generalization behavior of the model and its sensitivity to data scale.

\subsection{Experimental Setup}
\paragraph{Dstillation Setup.}
We distill two large-scale text-to-image generators, FLUX.1-12B~\cite{flux2024} and Qwen-Image-20B~\cite{wu2025qwenimagetechnicalreport}, into compact \name{} students. 
During distillation, the base parameters inherited from the teacher are kept frozen, while only the LoRA adapters are optimized.
Following Qwen-Image-Distill-LoRA \cite{DiffSynthStudio2025}, we extend the adapter placement beyond the standard attention projections. 
Specifically, LoRA modules are injected not only into the query, key, value, and output projections, but also into auxiliary projection layers and modality-specific MLPs across both visual and textual branches. 
This broader integration allows the student to more effectively capture cross-modal dependencies and retain generation fidelity despite its compact capacity.

\paragraph{Evaluation protocol.}
We conduct a comprehensive evaluation on 1024\textsuperscript{2} high-resolution image generation from three distinct prompt sets: 
(a) 10K captions from the COCO 2014 validation set \cite{lin2014microsoft}, 
(b) 3200 prompts from the HPSv2 benchmark \cite{wu2023human}, 
(c) 1120 prompts from OneIG-Bench \cite{chang2025oneig},
(d) 553 prompts from the GenEval benchmark \cite{ghosh2023geneval},
(e) 1065 prompts from the DPG-Bench \cite{hu2024ella},
and (f) 160 prompts from the LongText-Bench \cite{geng2025xomnireinforcementlearningmakes}.
For the COCO and HPSv2 sets, we report common metrics
including FID \cite{FID}, patch FID (pFID) \cite{lin2024sdxllightningprogressiveadversarialdiffusion}, CLIP similarity \cite{clip}, VQAScore \cite{vqascore}, and HPSv2.1 \cite{hpsv2}. 
On COCO prompts, FIDs are computed against real images, reflecting data alignment. 
On HPSv2, CLIP and VQAScore measure prompt alignment, while HPSv2 captures human preference alignment. 
For OneIG-Bench, GenEval, DPG-Bench, and LongText-Bench, we adopt their official evaluation protocols and report results based on their respective benchmark metrics.

\subsection{Main Results}

We compare \ours against other few-step student models distilled from the same teacher. 
For FLUX, we compare against: 
8-NFE Hyper-FLUX \cite{hypersd}, trained with consistency distillation (CD) and reward models (Re); 
8-NFE FLUX Turbo \cite{sauer2025adversarial}, based on GAN-like adversarial distillation;
$\pi$-Flow (FLUX) \cite{chen2025pi}, trained with policy-based imitation distillation ($\pi$-ID).
For Qwen-Image, we compare with the 4-NFE Qwen-Image Lighting based on variational score distillation (VSD) and $\pi$-Flow (Qwen). To further evaluate under extremely low-data conditions, we additionally implement a LoRA-based uniform timestep distillation approach that uniformly distill time steps using only 1  sample.
The results are shown in Table \ref{tab:fluxqwen} and \ref{tab:oneig}.
In Appendix A, we also explore the Qwen-Image-Edit model \cite{wu2025qwenimagetechnicalreport} in the image-editing domain.
% All quantitative results are presented in Table \ref{tab:fluxqwen} and \ref{tab:oneig}.

\paragraph{Overall Performance.}
As illustrated in Figure~\ref{fig:all_bench}, \ours exhibits performance curves that closely track those of the base models (Qwen-Image and Flux) across all benchmarks, indicating strong consistency under the accelerated setting.
The detailed quantitative summaries in Table~\ref{tab:fluxqwen} and Table~\ref{tab:oneig} confirm this trend: 
\ours achieves nearly the same generation quality as the 50-step base models while running \textbf{5$\times$ faster}.
Despite the aggressive acceleration, it maintains competitive FID, CLIP, and HPSv2 scores across all benchmarks, showing no clear degradation in visual fidelity or prompt alignment. 
Moreover, even when trained with only \textbf{1 sample} within less than one GPU hour, \ours delivers results comparable to other few-step distillation approaches that require large-scale datasets and heavy computation.
These results underscore the strong \textbf{efficiency–performance balance} achieved by our approach, validating that the proposed Slow-Fast LoRA framework can preserve high generation quality under \textit{minimal supervision and limited computational resources}.

\paragraph{Visual Fidelity.}
To further examine the trade-off between speed and quality, Figure~\ref{fig:compare_with_base_model} presents qualitative comparisons among our \ours models (8- and 10-step), the base teacher models (8-, 10-, and 50-step), and other 4-step distillation models such as Qwen-Image-Lightning and $\pi$-Flow, all under identical noise initialization. 
Even at only eight steps, \ours maintains the teacher’s global semantics and color composition with minimal loss of fidelity. 
Increasing the step count gradually restores fine textures and small structures, indicating that phase-aware LoRA adaptation preserves the denoising trajectory of teacher despite extreme acceleration. 
These observations align with our quantitative findings: the 5× faster model achieves nearly the \textbf{same quality as the teacher} while requiring only a fraction of the computation.

\subsection{Ablation Study}

We conduct comprehensive ablation studies to analyze the key factors that contribute to the effectiveness of the proposed Slow–Fast design. 
Unless otherwise stated, all experiments are evaluated on OneIG-Bench using Qwen-image as the teacher model.

\paragraph{The importance of Slow-Fast Design.}
To verify the effectiveness of the proposed phase-aware design, we divide the diffusion process into two distinct denoising stages and systematically vary the LoRA assignment strategy. 
Specifically, we experiment with five configurations to assess the role of Slow-LoRA and Fast-LoRA under different timestep allocations. 
We experiment with five configurations: (1) phase-aware Slow3–Fast5 setup (ours), (2) Slow3 + Base5, (3) Base3 + Fast5, (4) single LoRA at identical timesteps, and (5) single LoRA uniformly sampled across eight timesteps.

As summarized in Table \ref{tab:abla1}, our asymmetric Slow–Fast configuration achieves the \textbf{highest performance across all metrics}, demonstrating its superior balance between quality and efficiency.
This confirms that aligning LoRA updates with the semantic-to-refinement progression of denoising leads to more effective \textbf{knowledge transfer}. 
In this process, the model performs slow adaptation during early stages and fast refinement during later stages, achieving better specialization than uniform or single-expert alternatives.
Among all variants, the \textbf{Single (uniform)} setup performs the worst, confirming the necessity of phase-wise specialization.
Notably, we observe that the early-stage Slow-LoRA contributes more significantly to final image quality, underscoring the importance of coarse-to-fine adaptation in guiding generation.

% z

\begin{table}[h]
    \centering
    \caption{
    \textbf{Slowfast stage ablation study.}
    }
    
\vspace{-1em}
    \label{tab:abla1}
    \resizebox{\linewidth}{!}{%
        \begin{tabular}{lccccc}
        \toprule
        \textbf{Model} & \textbf{Alignment\textuparrow} & \textbf{Text\textuparrow} & \textbf{Diversity\textuparrow} & \textbf{Style\textuparrow}  & \textbf{Reasoning\textuparrow} \\
        \midrule
        Slow3–Fast5 
            & \textbf{0.849} & \textbf{0.614} & \textbf{0.152} & \textbf{0.396} & \textbf{0.284} \\
        Slow3 + Base5 
            & 0.805 & 0.567 & 0.123 & 0.368 & 0.255 \\
        Base3 + Fast5 
            & 0.747 & 0.521 & 0.125 & 0.372 & 0.243 \\
        Single (identical)
            & 0.702 & 0.453 & 0.110 & 0.342 & 0.218 \\
        Single (uniform)
            & 0.621 & 0.332 & 0.097 & 0.298 & 0.193 \\
        \bottomrule
        \end{tabular}
    }
    
\vspace{-2em}
\end{table}

\paragraph{If more samples help?}
To investigate the effect of data composition on LoRA adaptation, we first randomly select 1 text–image pairs from the Qwen-Image-Self-Generated-Dataset \cite{DiffSynthStudio_QwenImage_SelfGeneratedDataset_2025} dataset as the minimal training set. 
We then scale up the dataset while keeping the total number of training epochs fixed. 
As shown in Table~\ref{tab:abla2}, increasing the number of training samples from 1 to 10 and further to 100 does not lead to notable performance gains. Most metrics remain nearly unchanged, while the \textit{style} score even slightly declines, suggesting that simply enlarging the dataset without enhancing its diversity or phase alignment may weaken stylistic consistency.
These results indicate that for phase-aware LoRA adaptation, data quality and phase alignment are more crucial than scale, and that even a single well-chosen sample can achieve effective adaptation.
% \alex{Write observation here. Done}

\begin{table}[h]
    \centering
    \caption{
    \textbf{Training data ablation study}.
    % \alex{This table focus on data scale only.}
    }
    
\vspace{-1em}
    \label{tab:abla2}
    \resizebox{\linewidth}{!}{%
        \begin{tabular}{lccccc}
        \toprule
        \textbf{Model} & \textbf{Alignment\textuparrow} & \textbf{Text\textuparrow} & \textbf{Diversity\textuparrow} & \textbf{Style\textuparrow}  & \textbf{Reasoning\textuparrow} \\
        \midrule
        1 sample 
            & 0.868 & 0.734 & 0.160 & \textbf{0.421} & 0.303 \\
        10 samples
            & 0.874 & \textbf{0.758} & 0.163 & 0.414 & 0.296 \\
        100 samples
            & \textbf{0.876} & 0.753 & \textbf{0.165} & 0.418 & \textbf{0.306} \\
        \bottomrule
        \end{tabular}
    }
    
\vspace{-2em}
\end{table}

\paragraph{Timestep Ablation.}
In this experiment, we study the influence of the number of timesteps equipped with LoRA adapters while keeping the data selection and scale fixed at 1 text–image pairs. 
We progressively increase the number of timesteps on which LoRA modules are attached, thereby examining how the temporal coverage of adaptation affects generation quality. 
As shown in Table~\ref{tab:abla3}, with more LoRA-equipped timesteps, the overall performance steadily improves, particularly showing a clear gain in the \textit{text} metric.
This trend indicates that \textbf{broader temporal adaptation enables the model to capture more phase-specific denoising dynamics}, leading to better overall reconstruction quality.

\begin{table}[h]
    \centering
    \caption{\textbf{Timestep ablation study}.}
    
\vspace{-1em}
    \label{tab:abla3}
    \resizebox{\linewidth}{!}{%
        \begin{tabular}{lccccc}
        \toprule
        \textbf{Model} & \textbf{Alignment\textuparrow} & \textbf{Text\textuparrow} & \textbf{Diversity\textuparrow} & \textbf{Style\textuparrow}  & \textbf{Reasoning\textuparrow} \\
        \midrule
        Slow3 + Fast5 
            & 0.863 & 0.692 & 0.162 & 0.414 & 0.286 \\
        Slow5 + Fast5 
            & 0.868 & 0.734 & 0.160 & 0.421 & 0.303 \\
        Slow5 + Fast10 
            & \textbf{0.874} & \textbf{0.813} & \textbf{0.175} & \textbf{0.422} & \textbf{0.305} \\
        \bottomrule
        \end{tabular}
    }
    
\vspace{-1em}
\end{table}

\subsection{Discussion}

\paragraph{Generalization Ability of Different Single Training Samples.}
% \alex{List setting details. Done}
We conduct an ablation study using four distinct single-image settings to examine the generalization behavior of our framework under extreme data scarcity. 
Specifically, we first select \textit{in-distribution} samples from \cite{DiffSynthStudio_QwenImage_SelfGeneratedDataset_2025}, where the samples are generated from Qwen-image base model.
We select three sample with representing diverse semantic and structural characteristics: (1) \textbf{Fox}: an anthropomorphic fox with vibrant red fur and white facial features, (2) \textbf{Valley landscape}: a unique natural scene featuring a winding spiral valley, and (3) \textbf{Bookstore}: a text-rich storefront window densely filled with books and signage.
Additionally, an \textit{out-of-distribution (OOD)} real-world image, depicting a bustling city crosswalk filled with pedestrians moving in different directions on a clear day, is included to assess robustness beyond the training domain.
For completeness, we also experiment with an extreme setting where the model is trained purely on \textbf{Gaussian noise}, serving as a control to isolate the effect of meaningful visual content.
All models are trained on different samples using the same configuration to assess their generalization ability. 

\begin{figure}[t]
  \centering

   \includegraphics[width=\columnwidth]
   {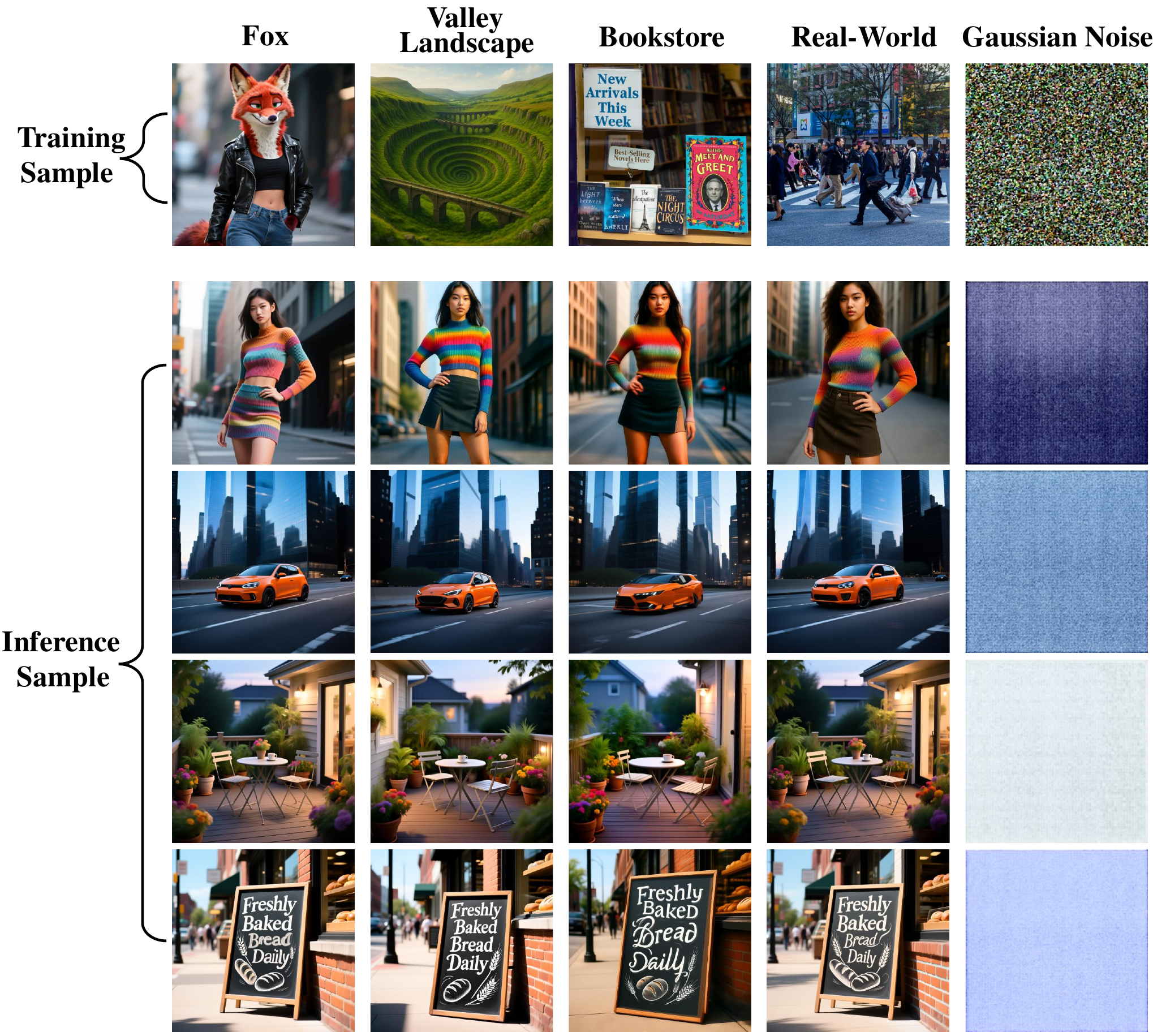}
   
\vspace{-1em}
    \caption{
    \textbf{Qualitative results from the one-sample training setting}. 
    Even trained on a single image, the model generalizes well to unseen prompts, producing coherent and detailed results across diverse scenes.}

   \label{fig:one_sample}
   
\vspace{-1em}
\end{figure}
\begin{table}[h]
    \centering
    \caption{\textbf{1 data sample study}. }
    % * indicates that the outputs are invalid and the metrics are omitted.}
    
\vspace{-1em}
    \label{tab:abla4}
    \resizebox{\linewidth}{!}{%
        \begin{tabular}{lccccc}
        \toprule
        \textbf{Model} & \textbf{Alignment\textuparrow} & \textbf{Text\textuparrow} & \textbf{Diversity\textuparrow} & \textbf{Style\textuparrow}  & \textbf{Reasoning\textuparrow} \\
        \midrule
        Fox 
            & \textbf{0.868} & 0.734 & \textbf{0.160} & \textbf{0.421} & \textbf{0.303} \\
        Valley Landscape 
            & 0.842 & 0.712 & 0.146 & 0.409 & 0.299 \\
        Bookstore (text-rich) 
            & 0.797 & \textbf{0.751} & 0.131 & 0.373 & 0.267 \\
        Real-World (OOD) 
            & 0.857 & 0.728 & 0.153 & 0.420 & 0.298 \\
        % Gaussian Noise 
            % & * & * & * & * & * \\
        \bottomrule
        \end{tabular}
    }
\end{table}

From Table~\ref{tab:abla4} and Fig.~\ref{fig:one_sample} we observe: \emph{i}. While the three single-sample settings exhibit noticeable stylistic differences, their quantitative performance remains relatively close.
\emph{ii}. The Bookstore (text-rich) model exhibits the weakest generalization, while the fox-trained model demonstrates the strongest. 
The fox-based LoRA yields consistent improvements across all evaluation metrics, and Figure~\ref{fig:one_sample} further illustrates that it produces images with better convergence and richer fine-grained details. 
However, the gap between them is modest.
\emph{iii}. Surprisingly, the Real-world image (OOD) model also achieves competitive results, with scores on most metrics only slightly lower than those of the Fox-based LoRA, suggesting that \textit{out-of-domain samples can still provide meaningful transferable cues for denoising adaptation}.
\emph{iv}. When trained solely on Gaussian noise, the model fails to produce any meaningful images, indicating that effective denoising requires exposure to data that resembles natural image distributions.

Overall, while different samples lead to consistent performance trends, the quantitative differences are minor. 
This shows that \ours is not particularly sensitive to which single image is used for training, and can still learn transferable denoising behaviors from extremely limited but coherent supervision.
\begin{figure}[t]
  \centering

   \includegraphics[width=0.48\textwidth]{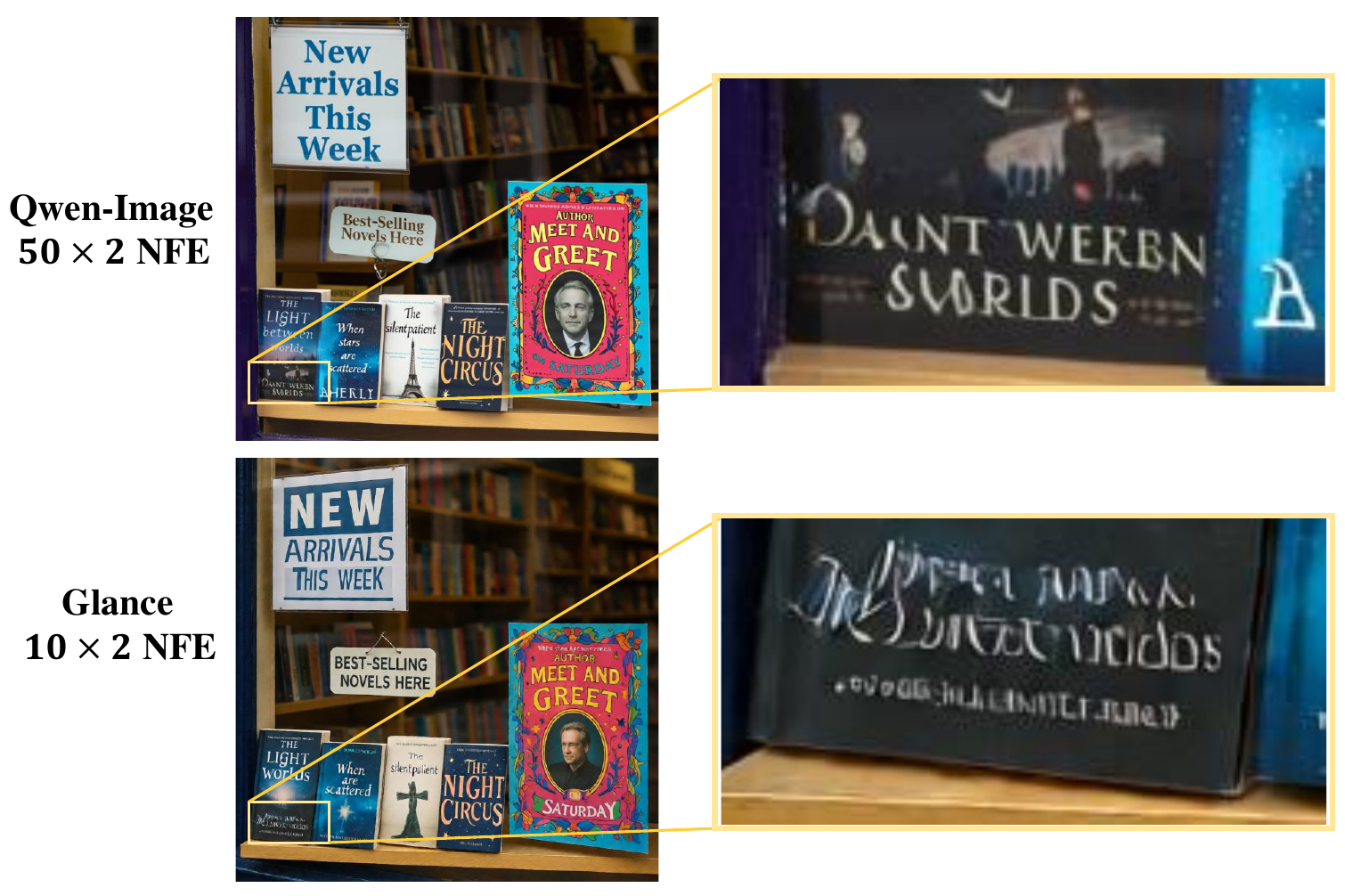}
   
\vspace{-1em}
   \caption{Text-render failure cases. \ours struggles on extremely small text, producing blurred or distorted characters.}
   \label{fig:failure}
   
\vspace{-2em}
\end{figure}

\paragraph{Failure Analysis.}
Although \ours achieves performance curves that closely track those of the base models across all evaluated benchmarks, we identify a consistent weakness in text rendering quality. As reported in Table~\ref{tab:oneig} and Fig.~\ref{fig:all_bench}, \ours shows clear deficits in both text rendering, falling behind Qwen-Image and FLUX by 0.154 and 0.228 on the Text-Render metric, and by 14.7 and 13.0 points on LongText-Bench, respectively.
To better understand this discrepancy, we conducted a detailed inspection of the generated samples. As illustrated in Fig.~\ref{fig:failure}, failure cases predominantly occur in images containing extremely dense or very small text, where the model struggles to preserve sharp character boundaries, often producing blurred strokes or local artifacts. In contrast, when the text is shorter and occupies a larger spatial extent in the image, \ours is able to reproduce it faithfully.

These suggest that high-frequency textual details are harder for the student to capture than general visual content, Such fine-grained details require precise spatial alignment and are harder to capture during distillation, especially under few-step constraints.

\section{Conclusion}
We present \ours, a lightweight distillation framework that accelerates diffusion inference through a phase-aware Slow–Fast design. 
The well studied LoRA adapters distinct denoising phases to efficiently capture both global semantics and local refinements. 
\ours enables high-quality generation with only eight steps, achieving an \textbf{5$\times$ speed-up} over the base model.
Despite being trained with as few as one image and a few GPU hours, \ours maintains comparable visual fidelity and exhibits strong generalization to unseen prompts. 
These results highlight that data- and compute-efficient distillation can retain the expressive capacity of large diffusion models without sacrificing quality. 
We believe \ours serve as a strong candidate for accelerating large-scale diffusion models, particularly in data-scarce applications.
%\alex{Check if last sentence correct.} \rui{reorg}
% Future work will extend this framework to broader architectures and modalities, and explore its adaptation to special domains (e.g., medical, artistic, or scientific imagery), aiming to make step-efficient diffusion a general paradigm for \textbf{scalable generative modeling}.

{
    \small
    \bibliographystyle{utils/ieeenat_fullname}
    \bibliography{main}
}

\clearpage
\maketitlesupplementary

\appendix

\addtocontents{toc}{\protect\setcounter{tocdepth}{2}}

\tableofcontents

\section{Qwen-Image-Edit task}
Beyond the text-to-image setting, a natural question is whether our phase-aware distillation strategy can transfer to other generative tasks. Motivated by this curiosity, we further evaluate our approach on image-editing using the Qwen-Image-Edit model. 
\subsection{Training Setup}
\begin{figure}[t]
  \centering
   \includegraphics[width=0.48\textwidth]{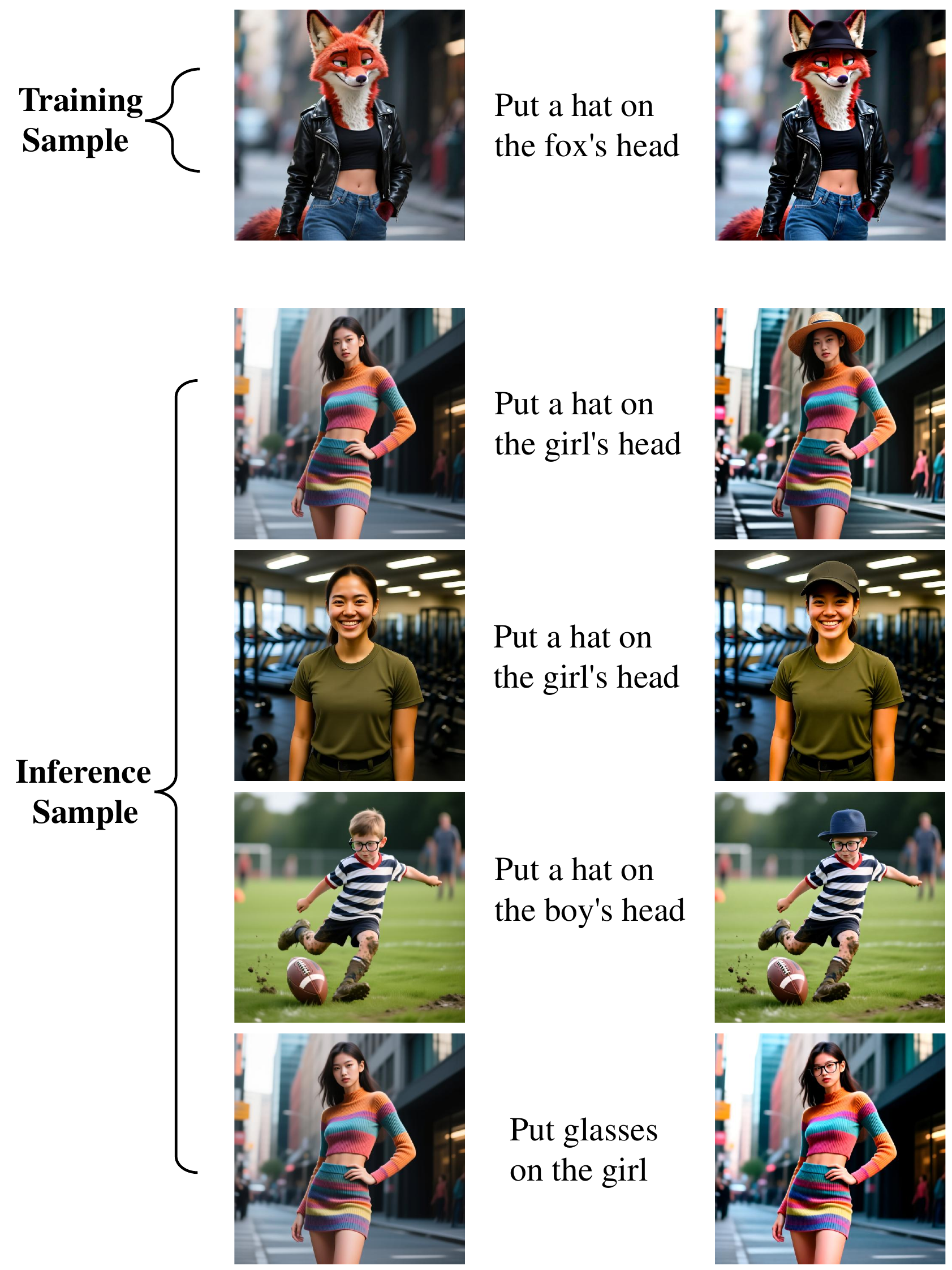}
   
   \caption{Training and inference examples for the one-shot Qwen-Image-Edit adaptation.}
   \label{fig:appendix_edit}
   
\end{figure}

Our training configuration closely follows the setup used for Qwen-Image experiments. We train a new pair of Slow-LoRA and Fast-LoRA adapters on the base Qwen-Image-Edit model, while adjusting the learning rate to 2e-4 to account for the different task dynamics.
Following the one-shot paradigm, we train the LoRA experts on a single image, shown in Fig.~\ref{fig:appendix_edit}, with the editing instruction \textit{“Put a hat on the woman's head.”}

\subsection{Results and Analysis}

Despite being trained on only one image, our method exhibits surprisingly strong generalization on the Qwen-Image-Edit task. Across a broad set of test samples, the edited images consistently follow the prompt: the model accurately places a hat on the target person while leaving the rest of the scene intact. This ability to preserve non-edited regions demonstrates that the Slow- and Fast-LoRA experts successfully adapt the model’s denoising phases without degrading spatial consistency.

\textbf{Interestingly, the model does not simply copy the hat from the training image.} Instead, it generates varied, context-appropriate hats for different individuals—evidence that the distilled LoRA experts capture high-level editing semantics rather than memorizing pixel-level details.

To further test the flexibility of our approach, we modify the instruction to \textit{“Put glasses on the girl.”} Even with this completely new editing concept, the model performs accurately and robustly, inserting realistic glasses while maintaining the appearance and identity of the original subject. The strong performance under unseen editing instructions highlights the task-agnostic nature of our distillation strategy.

Overall, these findings suggest that the proposed phase-aware LoRA acceleration is not limited to text-to-image synthesis; it naturally extends to other generative domains with minimal adaptation. In future work, we plan to explore broader applications, including controllable generation, inpainting, and video editing, to further uncover the potential of this lightweight and generalizable distillation framework.

\section{Remote Sensing Domain Speedup}
\subsection{Zero-Shot Behavior}
One-shot acceleration in \ours produced surprisingly strong results on common natural images, which motivated us to explore domains where data availability is inherently limited. Remote sensing imagery is a representative low-resource domain: its acquisition is expensive, often restricted by geographic or institutional constraints, and large-scale datasets are difficult to obtain. This unique setting raises a natural question—\textit{can phase-aware acceleration with Glance generalize to such domains without requiring extensive retraining?}

To investigate this, we directly applied a \ours model trained only on random natural images to remote sensing image generation. As shown in Fig.~\ref{fig:appendix_rs}, the model produces visually plausible outputs and maintains high-fidelity structures. However, the generated perspectives resemble natural images rather than true overhead remote-sensing viewpoints. This observation indicates that while \ours generalizes appearance, zero-shot adaptation is insufficient to inject domain-specific geometric priors.

\subsection{One-Shot Slow-Fast LoRA Adaptation}

\begin{figure}[t]
  \centering
   \includegraphics[width=0.48\textwidth]{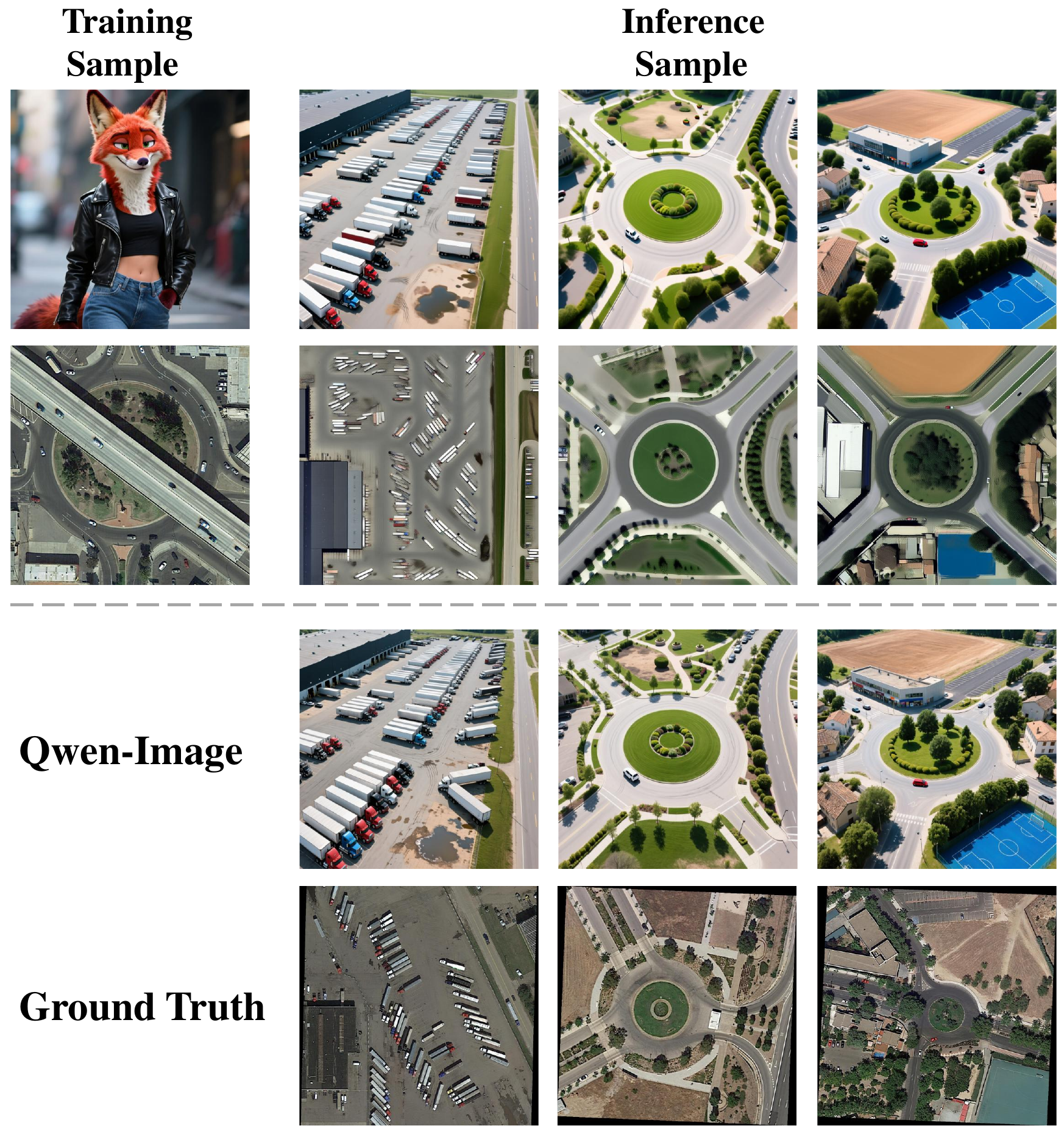}
   
   \caption{\textbf{Qualitative results from the one-sample training setting.} After observing only a single remote sensing example, \ours adapts effectively and begins generating images that exhibit correct aerial viewpoints and characteristics consistent with real remote sensing imagery.}
   \label{fig:appendix_rs}
   
\end{figure}

To address this, we performed a targeted adaptation by training the Slow-LoRA and Fast-LoRA experts using only a single remote sensing sample ((Fig.~\ref{fig:appendix_rs}). After this minimal fine-tuning, the adapted \ours model exhibits a remarkable change in behavior. When tested on unseen prompts, it consistently produces images with proper aerial viewpoints and structural layouts characteristic of remote sensing photography. This demonstrates that the model successfully captures domain-specific priors from just one example, while maintaining the acceleration benefits provided by the phase-aware LoRA design.
The strong generalization achieved with minimal supervision highlights the effectiveness of Slow–Fast LoRA in injecting domain-specific knowledge without sacrificing speed or requiring large-scale datasets.

For reference, the single training example used for adaptation is shown in  
Fig.~\ref{fig:appendix_rs}. Its detailed description is as follows:
\textit{A high-resolution overhead satellite image of a large urban roundabout intersected by an elevated roadway.
A wide overpass runs diagonally across the scene from bottom-left to top-right, casting a long shadow onto the circular traffic island below.
Beneath the overpass lies a landscaped roundabout with grass, shrubs, and groups of small trees arranged in patches.
At the center of the roundabout is a paved triangular plaza with a statue or vertical monument casting a distinct shadow.
Multiple cars of various colors—blue, white, black, and red—drive along the circular lanes around the roundabout, following curved road markings.
Surrounding the intersection are urban buildings, parking lots, and paved sidewalks.
The lighting suggests clear weather and midday sun, creating sharp shadows from vehicles, trees, and the elevated bridge.}

\section{More Implementation details}
\subsection{Glance (Qwen-Image)}

For Qwen-Image, 
in Slow-LoRA and Fast-LoRA,
we apply LoRA to a broad set of multi-modal projection and modulation layers to ensure effective low-rank adaptation. Specifically, LoRA adapters are injected into the \textit{to\_q}, \textit{to\_k}, \textit{to\_v}, \textit{add\_q\_proj}, \textit{add\_k\_proj}, \textit{add\_v\_proj}, and \textit{to\_out.0} modules of the MM-DiT blocks. In addition, LoRA is also placed on the multimodal MLP and modulation pathways, including \textit{to\_add\_out}, \textit{img\_mlp.net.2}, \textit{img\_mod.1}, \textit{txt\_mlp.net.2}, and \textit{txt\_mod.1}. 

We set the LoRA rank and scaling parameter to $r = 32$ and $\alpha = 128$, respectively, and follow the Gaussian initialization strategy for all LoRA weight matrices.

During training, we use a learning rate of $3 \times 10^{-4}$ with a constant schedule. The optimizer is AdamW with $\beta_{1} = 0.9$, $\beta_{2} = 0.999$, weight decay of $10^{-2}$, and $\epsilon = 10^{-8}$. Training is performed with a global batch size of $1$, mixed-precision (\texttt{bf16}), and gradient clipping of $1.0$. We train for a total of $60$ steps, which corresponds to effectively training the single data sample for $60$ epochs, and enable 8-bit Adam and quantized weight loading to reduce memory footprint. Both image and text embeddings are precomputed to accelerate training.

\subsection{Glance (FLUX)}
For FLUX,
in Slow-LoRA and Fast-LoRA,
we employ LoRA on the \textit{to\_q}, 
\textit{to\_k}, 
\textit{to\_v} and \textit{to\_out.0} modules of the MM-DiT.

We set the LoRA rank and scaling parameter to $r = 16$ and $\alpha = 64$, respectively, and follow the Gaussian initialization strategy for all LoRA weight matrices.

During training, we use a learning rate of $5 \times 10^{-4}$ with a constant schedule. The optimizer is AdamW with $\beta_{1} = 0.9$, $\beta_{2} = 0.999$, weight decay of $10^{-2}$, and $\epsilon = 10^{-8}$. Training is performed with a global batch size of $1$, mixed-precision (\texttt{bf16}), and gradient clipping of $1.0$. We train for a total of $60$ steps, which corresponds to effectively training the single data sample for $60$ epochs, and enable 8-bit Adam and quantized weight loading to reduce memory footprint.

\section{More Qualitative Results}
\subsection{Glance (Qwen-Image)}

\begin{figure*}[t]
  \centering
   \includegraphics[width=0.95\textwidth]{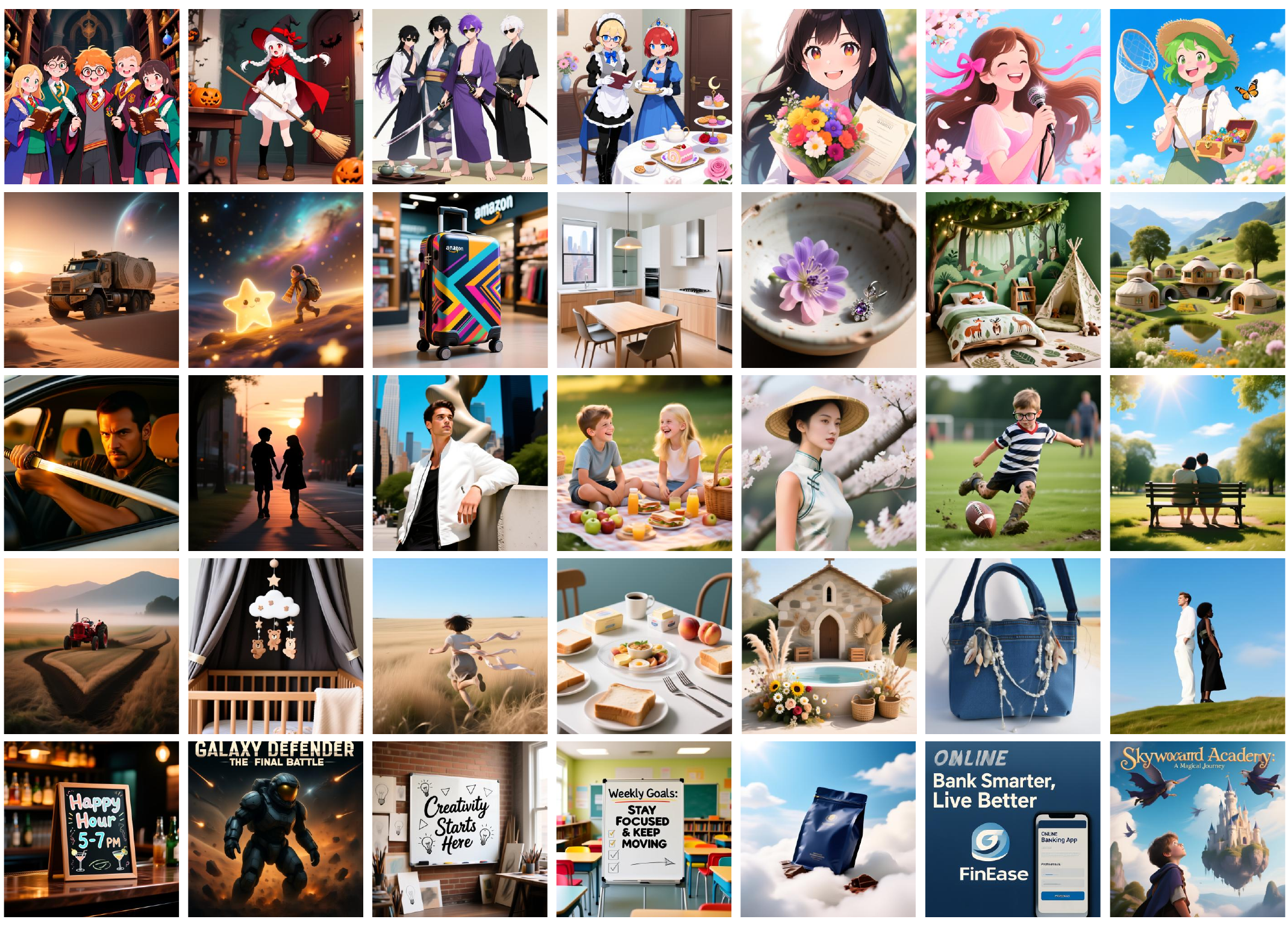}
   
   \caption{
    An uncurated random batch from the OneIG-Bench prompt set.
   }
   \label{fig:appendix1}
   
\end{figure*}

\begin{figure*}[t]
  \centering
   \includegraphics[width=0.95\textwidth]{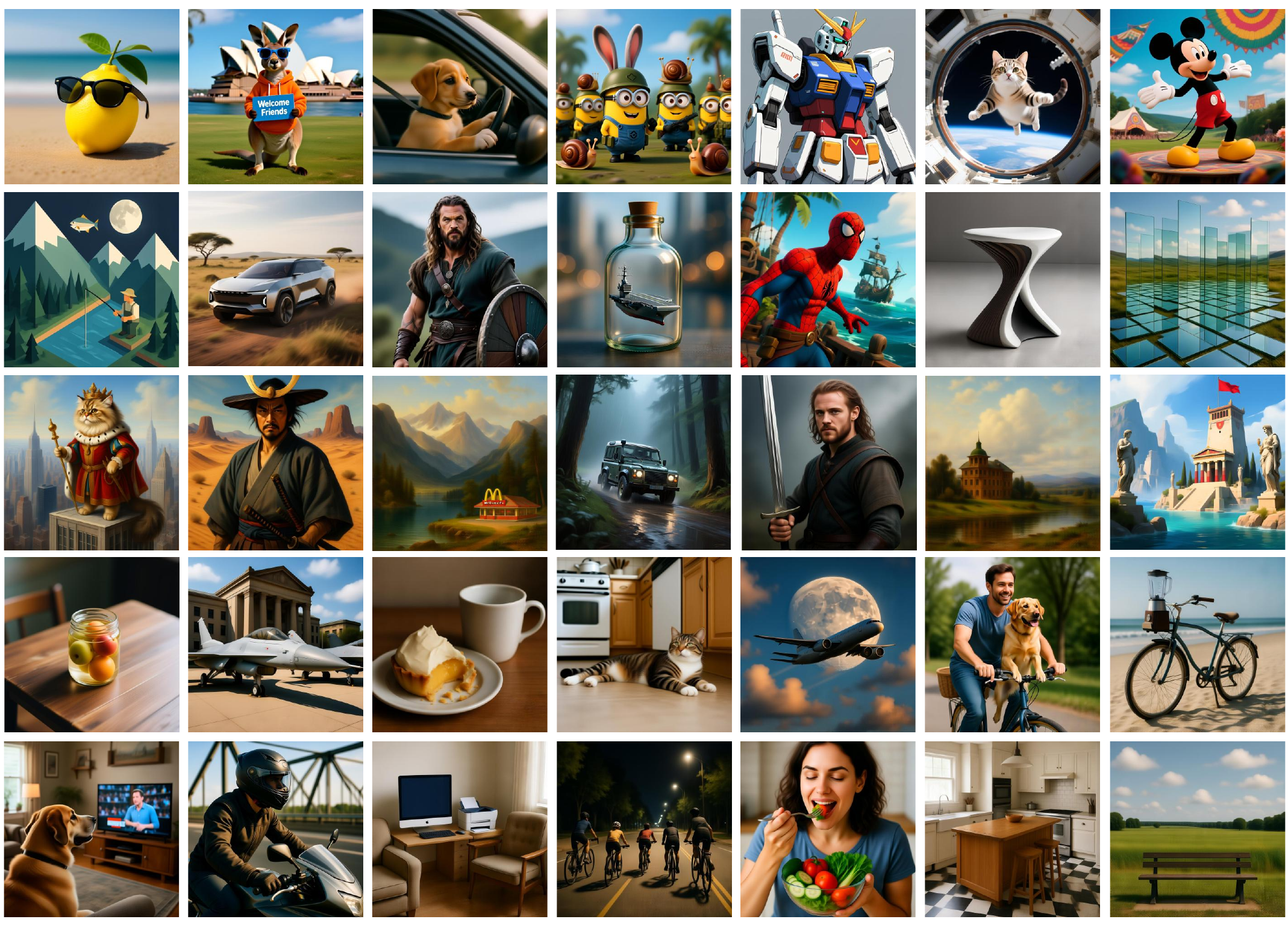}
   
   \caption{
    An uncurated random batch from the HPSv2 prompt set.
   }
   \label{fig:appendix2}
   
\end{figure*}

We show additional uncurated results of \ours (Qwen) in Fig.~\ref{fig:appendix1} and~\ref{fig:appendix2}.


\subsection{Glance (FLUX)}

\begin{figure*}[t]
  \centering
   \includegraphics[width=\textwidth]{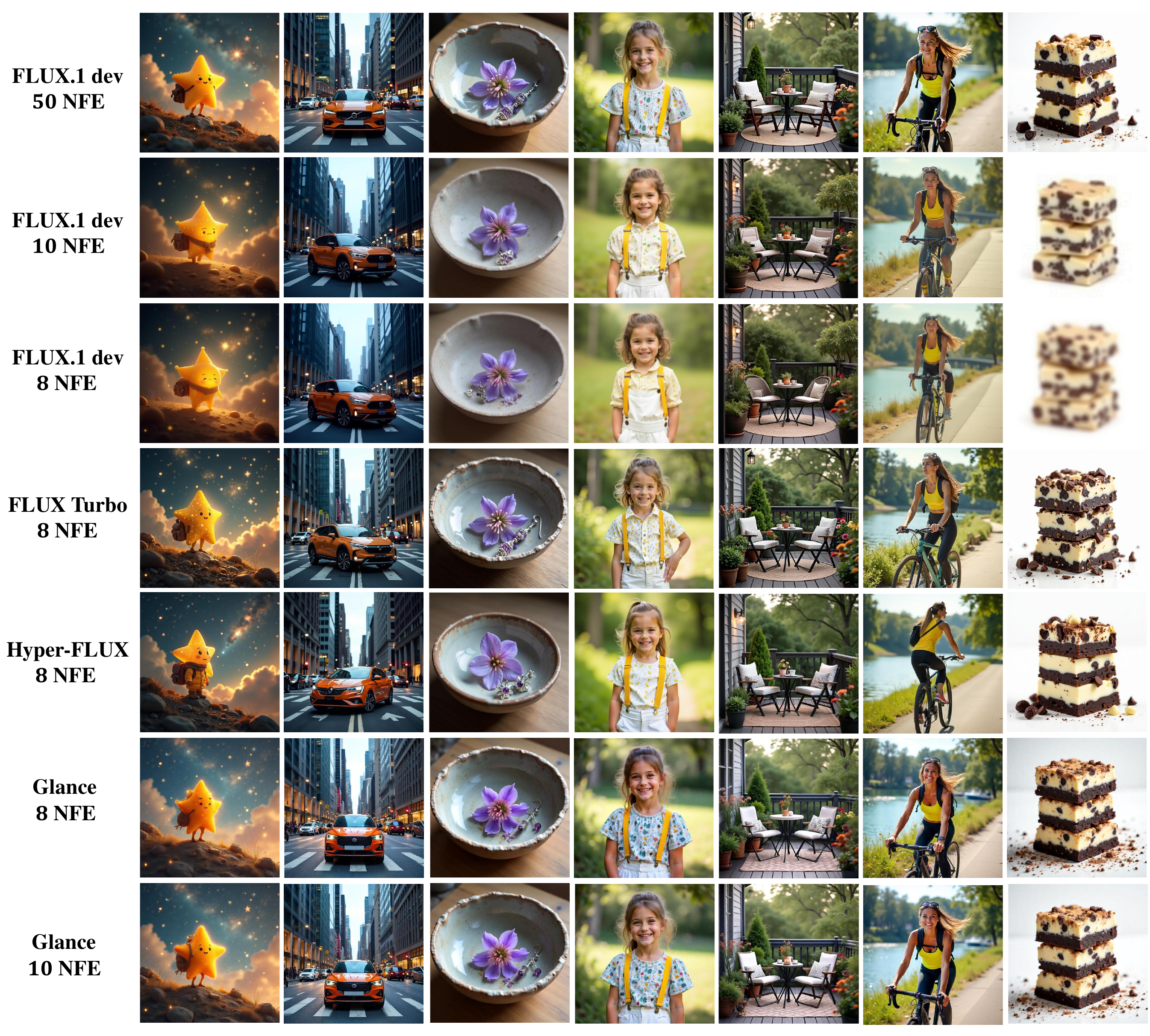}

   \caption{
    \textbf{Visual comparison of different Slow–Fast configurations.}
    All images are generated from the same initial noise using the 50-step base model, our 8/10-step students, and other few-step models.
    \name{} preserves semantic fidelity under strong acceleration, while additional steps progressively enhance fine details.
   }
   \label{fig:appendix3}
   
\end{figure*}

\begin{figure*}[t]
  \centering
   \includegraphics[width=0.95\textwidth]{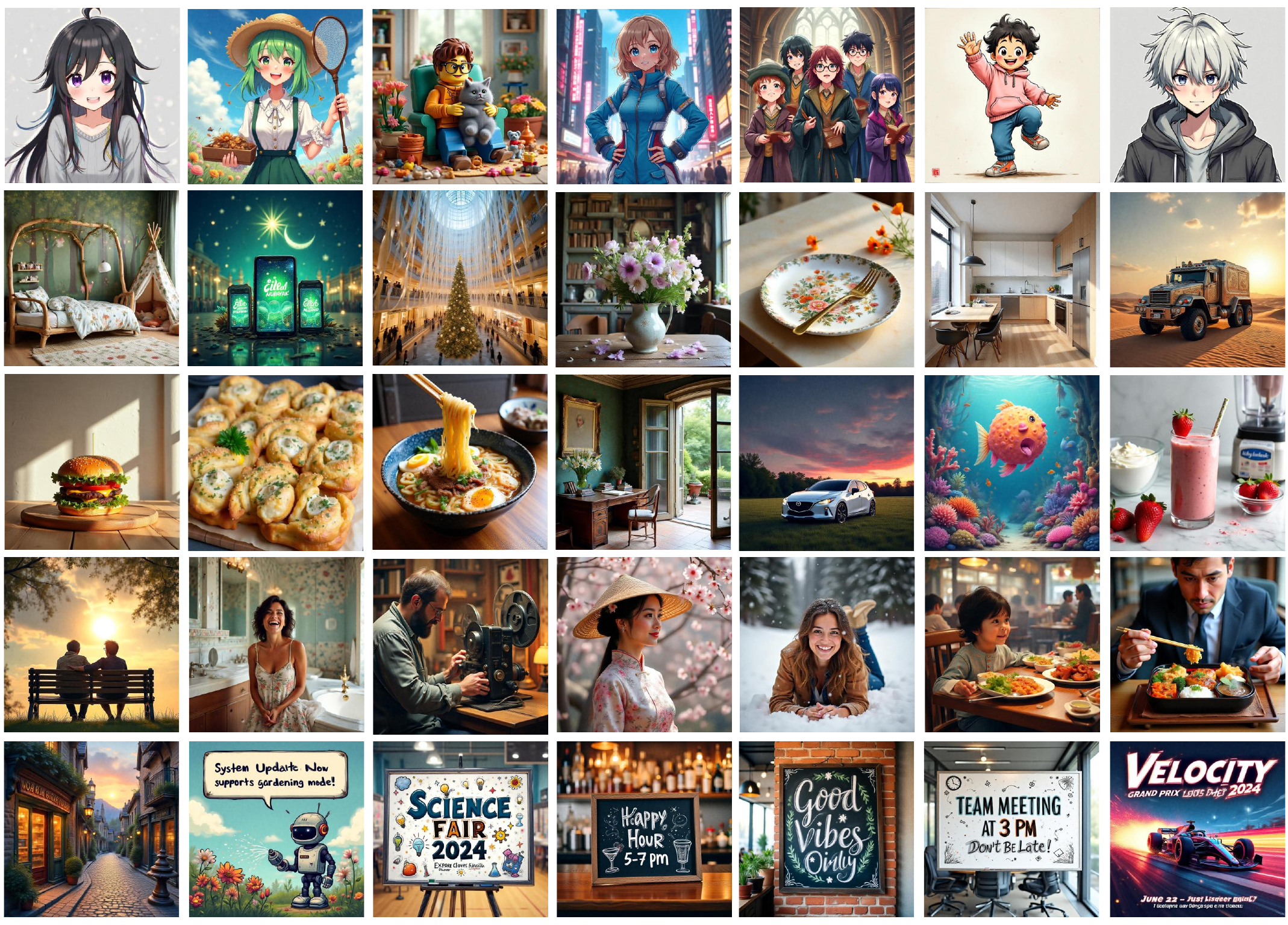}
   
   \caption{
    An uncurated random batch from the OneIG-Bench prompt set.
   }
   \label{fig:appendix4}
   
\end{figure*}

\begin{figure*}[t]
  \centering
   \includegraphics[width=0.95\textwidth]{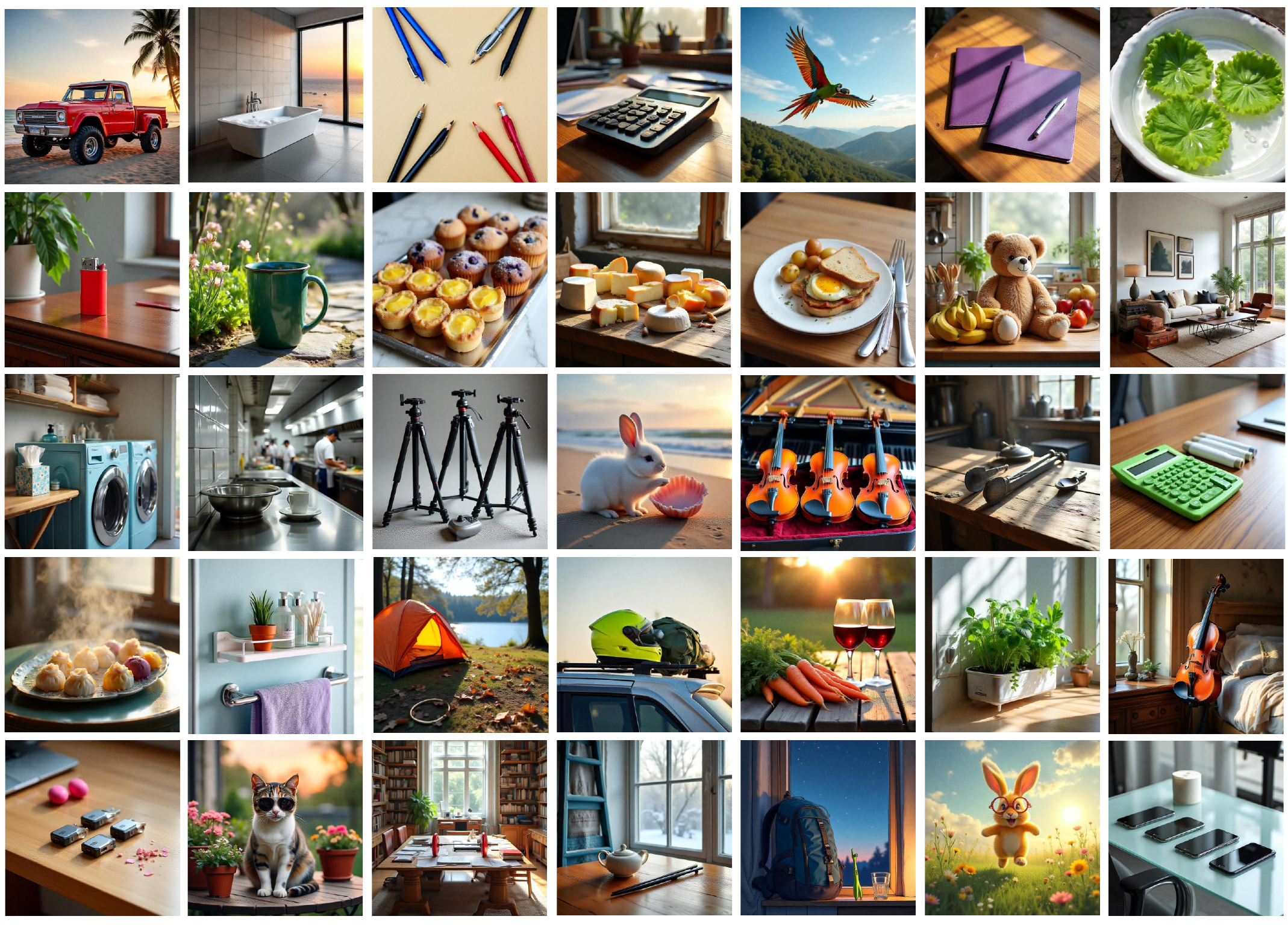}
   
   \caption{
    An uncurated random batch from the DPG prompt set.
   }
   \label{fig:appendix5}
   
\end{figure*}

We show additional uncurated results of \ours (FLUX) in Fig.~\ref{fig:appendix3}, Fig.~\ref{fig:appendix4} and~\ref{fig:appendix5}.

\section{Future Work}

Although our phase-aware acceleration framework already achieves strong performance with extremely lightweight adaptation, several promising directions remain open for exploration.

\paragraph{Dynamic Expert Switching Beyond Hard SNR Thresholds.}
In the current design, \ours employs a hard switch between Slow-LoRA and Fast-LoRA based solely on the SNR-based phase boundary. While effective, this strategy does not account for prompt-dependent difficulty. A more adaptive alternative is to dynamically adjust the switching point according to the complexity of the generation task. For challenging prompts that involve intricate structures or fine-grained semantics, the model could remain longer in the slow-denoising phase to preserve fidelity. Conversely, for simpler prompts, the model could transition earlier into the fast phase to maximize speedup. Learning such prompt-aware switching policies represents an exciting opportunity for further reducing inference cost while maintaining high visual quality.

\paragraph{Toward Zero-Shot Diffusion Distillation.}
Our method demonstrates that phase-aware LoRA with only a single training sample is already sufficient to capture strong domain priors and generalize effectively. A natural next step is to explore whether \ours can be extended to a complete zero-shot distillation setting. This would involve leveraging intrinsic diffusion priors, self-consistency constraints, or synthetic trajectories generated by the model itself, enabling fully data-free adaptation. Achieving robust zero-shot distillation would further push the boundary of efficient diffusion acceleration and enable deployment in domains where even a single reference example is unavailable.

\end{document}